\documentclass[12pt]{article}
\makeatletter                 
\usepackage{hyperref}
\usepackage{enumitem}
\usepackage{longtable}
\usepackage{soul}
\usepackage{xcolor}
\usepackage{amssymb}
\usepackage{tabularx} 
\usepackage{booktabs} 
\usepackage[]{natbib}
\usepackage{amsmath,amsfonts,bm}
\usepackage[parfill]{parskip}
\usepackage{amsthm}
\usepackage[toc,page]{appendix}

\usepackage{setspace}
\usepackage[pagewise]{lineno}
\usepackage{ragged2e}
\usepackage{booktabs} 
\usepackage{array}    
\usepackage{multirow} 
\usepackage{microtype}
\usepackage{graphicx}
\usepackage{booktabs} 
\usepackage{multirow} 
\usepackage{wrapfig}
\usepackage{comment}
\usepackage{caption}
\usepackage{bbm}
\usepackage{subcaption}
\usepackage{enumitem}
\setitemize{noitemsep,topsep=0pt,parsep=0pt,partopsep=0pt}
\usepackage{rotating}
\usepackage{makecell} 
\usepackage{algorithm}
\usepackage{chngcntr} 
\usepackage[noend]{algpseudocode}
\algrenewcommand\textproc{}
\algdef{SE}[SUBALG]{Indent}{EndIndent}{}{\algorithmicend\ }%
\algtext*{Indent}
\algtext*{EndIndent}

\usepackage{threeparttable}

\numberwithin{equation}{section}

\usepackage{algorithm}
\hypersetup{
	unicode=false,
	pdftoolbar=true,
	pdfmenubar=true,
	pdffitwindow=false,     
	pdfstartview={FitH},    
	pdfkeywords={}, 
	pdfnewwindow=true,      
	colorlinks=true,       
	linkcolor=cyan,          
	citecolor=blue,        
	filecolor=cyan,      
	urlcolor=cyan           
}

\usepackage[utf8]{inputenc}

\newcommand{\neighbor}[1]%
{\overline{#1}}

\newcommand{\blind}{1}
\usepackage{colortbl}
\definecolor{lightblue}{rgb}{0.8, 0.9, 1}
\usepackage{mathtools}
\graphicspath{{fig/}}

\begin{document}

\def\spacingset#1{\renewcommand{\baselinestretch}%
{#1}\small\normalsize} \spacingset{1.4}

\if1\blind
{
  \title{ \bf Fin-R1: A Large Language Model for Financial Reasoning through Reinforcement Learning}
   \author{Zhaowei Liu$^{1}$, Xin Guo$^{1}$,   Zhi Yang$^{1}$, Fangqi Lou$^{1}$, Lingfeng Zeng$^{1}$, Jinyi Niu$^{2}$, \\ 
    Mengping Li$^{1}$, Qi Qi$^{1}$, Zhiqiang Liu$^{1}$,  Yiyang Han$^{3}$, Dongpo Cheng$^{4}$, \\
    Ronghao Chen$^{5}$, Huacan Wang$^{5}$, 
   Xingdong Feng$^{1}$, Huixia Judy Wang$^{6}$, \\ Chengchun Shi$^{7*}$, and Liwen Zhang$^{1,8}$\thanks{Address for correspondence: Liwen Zhang (\texttt{zhang.liwen@shufe.edu.cn}), Chengchun Shi (\texttt{c.shi7@lse.ac.uk})} \\ 
  \smallskip\\
 \small  $^1$School of Statistics and Data Science, Shanghai University of Finance and Economics, China
 \\
\small  $^2$School of Mathematical Sciences, Fudan University, China
\\
\small  $^3$School of Economics,  Shanghai University of Finance and Economics, China
\\
\small  $^4$AI Finance Development and Service Center, Shanghai University of Finance and Economics, China
\\
\small $^5$QuantaAlpha, China
\\
\small $^6$Department of Statistics,  Rice University, USA
\\
\small $^7$Department of Statistics, London School of Economics and Political Science, UK
\\
\small $^8$Qinghai Provincial Key Laboratory of Big Data in Finance and Artificial  Intelligence \\\small Application Technology, 
Qinghai Institute of Technology, China
}
\date{}
  \maketitle
} \fi

	
\vspace{-1.5em}
\begin{abstract}
In recent years, general-purpose large language models (LLMs) such as GPT, Gemini, Claude, and DeepSeek have advanced at an unprecedented pace. Despite these achievements, their application to finance remains challenging, due to fragmented data sources,    reasoning processes, and weak transferability to business applications.
In response, we introduce  \textbf{Fin-R1}, a reasoning LLM designed for financial scenarios. With a compact size of 7 billion parameters, Fin-R1 reduces deployment costs while  addressing the aforementioned challenges. Its development follows a two-stage pipeline.
First, we construct \textbf{Fin-R1-Data}, a high-quality financial dataset consisting of 60,091 chain-of-thought (CoT) samples,  distilled and filtered from multiple authoritative benchmarks to ensure consistency and reliability. 
Second, we train Fin-R1 using Fin-R1-Data through supervised fine-tuning (SFT), followed by reinforcement learning (RL). This stage substantially improves the model’s ability to solve complex financial reasoning tasks, yielding outputs that are both accurate and interpretable. Despite its relatively small parameter scale, Fin-R1 achieves competitive empirical performance across established financial benchmarks and demonstrates practical utility in compliance checking and robo-advisory. Our code is publicly available at \url{https://github.com/SUFE-AIFLM-Lab/Fin-R1}, and has already attracted over 700 stars.
\end{abstract}


\spacingset{1.7} 

\section{Introduction}\label{sec-intro}
Large language models (LLMs) are highly sophisticated neural networks that can understand and generate human language. 
In recent years, they have achieved groundbreaking progress in natural language processing, making an important step toward artificial general intelligence~\citep[AGI,][]{touvron2023llama,team2024gemini,naveed2025comprehensive}. 
At the core of LLMs lies a two-stage training paradigm. The first stage, \textit{pre-training} \citep{bai2023qwen,anil2023palm}, leverages the transformer architecture~\citep{vaswani2017attention} to enable LLMs to learn linguistic patterns, world knowledge, and basic reasoning capacities from vast amounts of unlabeled text. However, pre-trained LLMs remain limited in their ability to handle complex reasoning tasks. This limitation has led to the second stage, \textit{post-training}, where customized algorithms have been developed to enhance reasoning ability and improve output quality. These algorithms include supervised fine-tuning \citep[SFT,][]{ouyang2022training} to inject reasoning knowledge, reinforcement learning from human feedback \citep[RLHF,][]{christiano2017deep} to improve output quality, and reinforcement learning with verifiable rewards \citep[RLVR,][]{lambert2024tulu} to optimize the model's reasoning capabilities through objectively verifiable signals. 
Although post-training is conducted on small-scale data and constitutes less than 1\% of the total training computation~\citep{Lai_2025_Survey}, it plays a pivotal role in enhancing reasoning performance, task alignment, and response reliability, enabling LLMs to handle various complex tasks at near-human levels.

Building on these post-training algorithms, a new generation of general-purpose, reasoning-oriented LLMs has emerged, including OpenAI’s o1~\citep{chatgpt} and o1-like models such as QwQ~\citep{qwen} and Marco-o1~\citep{zhao2024marco}. These models explicitly incorporate an exploration–reflection–iteration (ERI) mechanism to enhance reasoning.  
In ERI, the model first explores multiple candidate reasoning traces, then verifies them through self-consistency or verifiable signals, and finally iterates to refine the reasoning process until convergence. 
This mechanism 
has led to significant improvements in chain-of-thought \citep[CoT,][]{wei2022chain}\footnote{CoT refers to a sequence of intermediate reasoning steps that can be elicited either through carefully designed few-shot examples or through some simple magical prompts (e.g., ``Let us think step by step'').} reasoning 
on complex tasks, with performance often reaching the level of human experts. 
It is particularly useful in mathematical and logical problems that require a deep understanding of the problems themselves, step-by-step reasoning, and precise solutions.
However, our empirical investigation reveals that such general-purpose, reasoning-oriented LLMs become limited when applied to the financial domain (see Section~\ref{s3}). These limitations 
stem primarily from three sources: 
\begin{enumerate}[leftmargin=*]
    \item Financial data are highly fragmented and lack a unified structure \citep{wang2024quantagent,guo2025fineval,li2024alphafin,dong2024fnspid}
, making 
the integration of knowledge extremely difficult. Specifically, information such as contractual terms, regulatory requirements, macroeconomic indicators, and market signals is often dispersed across heterogeneous sources, and can be inconsistent or even mutually contradictory. This not only increases the cost of data preprocessing but also explains why high-quality CoT data remain scarce in finance. As financial reasoning relies on the integration of economic, legal, and quantitative logic, the shortage of high-quality data limits the capability of existing models to reason coherently in finance.
\item Most existing LLMs still operate as “black boxes,” 
with reasoning processes that remain intransparent: we only observe the final outputs, but not the underlying reasoning paths \citep{wang2023alpha,zhao2024explainability,tong2024ploutos}. Such lack of transparency conflicts with the regulatory and compliance requirements of the financial sector where traceability and explainability are essential, restricting their deployment in practice.
\item In high-stakes financial applications, existing models often suffer from weak transferability and generalization, which makes their outputs unreliable~\citep{yu2024fincon,fatouros2024can,zhou2023universalner}. 
Specifically, financial tasks involve evolving environments, which challenge the ability of existing models to generalize and adapt across scenarios. For instance, models trained solely via SFT 
rely heavily on memorizing previously seen examples rather than performing reasoning grounded in financial logic \citep{zhang2025policy}. This largely limits their effectiveness in applications such as credit assessment and risk pricing.
\end{enumerate}

\begin{figure}[t]
    \centering 
    \begin{subfigure}[b]{\textwidth} 
        \centering
        \includegraphics[width=0.6\textwidth]{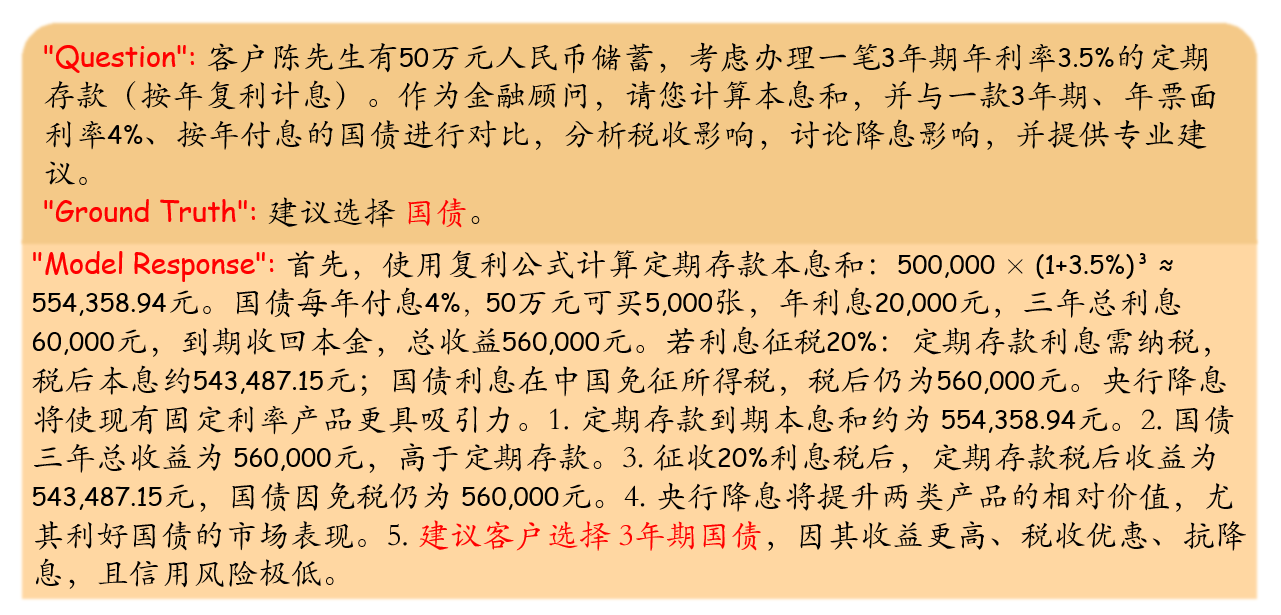} 
        \caption{Chinese interaction example}
        \label{demo1}
    \end{subfigure}
    \begin{subfigure}[b]{\textwidth}
        \centering
        \includegraphics[width=0.6\textwidth]{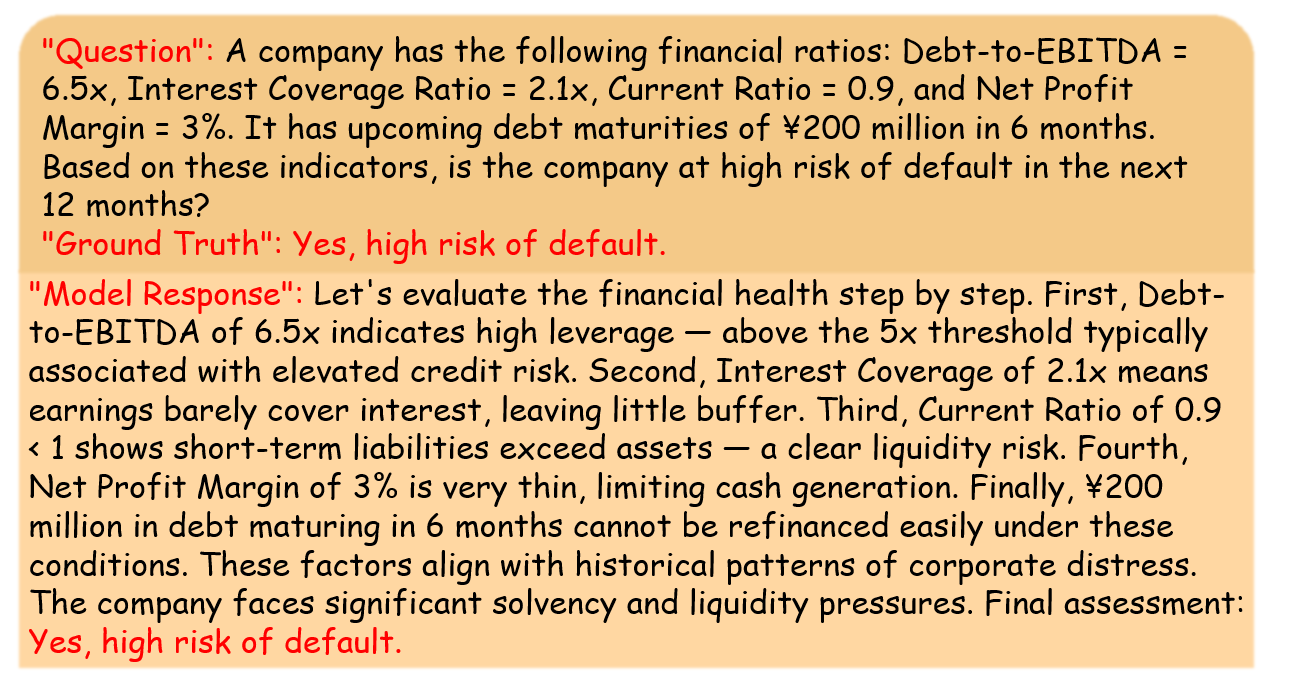}
        \caption{English interaction example}
        \label{demo2}
    \end{subfigure}
    \caption{Specific interaction examples based on Fin-R1 are provided, where Figure~\ref{demo1} and Figure~\ref{demo2} show QA related to government bonds in a Chinese context and QA related to debt risk in an English context, respectively. The final answers and ground truth in all model outputs are marked in the same color.}
    \label{fig:total}
\end{figure}

\textbf{Our contribution}. To overcome these limitations, we propose a two-stage framework for training domain-specific reasoning LLMs tailored to financial applications. 
The proposed framework consists of a \textit{\ul{data construction}} stage and a \textit{\ul{model training}} stage, and the overall pipeline is shown in Figure~\ref{fig:2stage}. Specifically: 
\begin{sidewaysfigure}[htbp]  
  \centering
  \includegraphics[width=\textwidth]{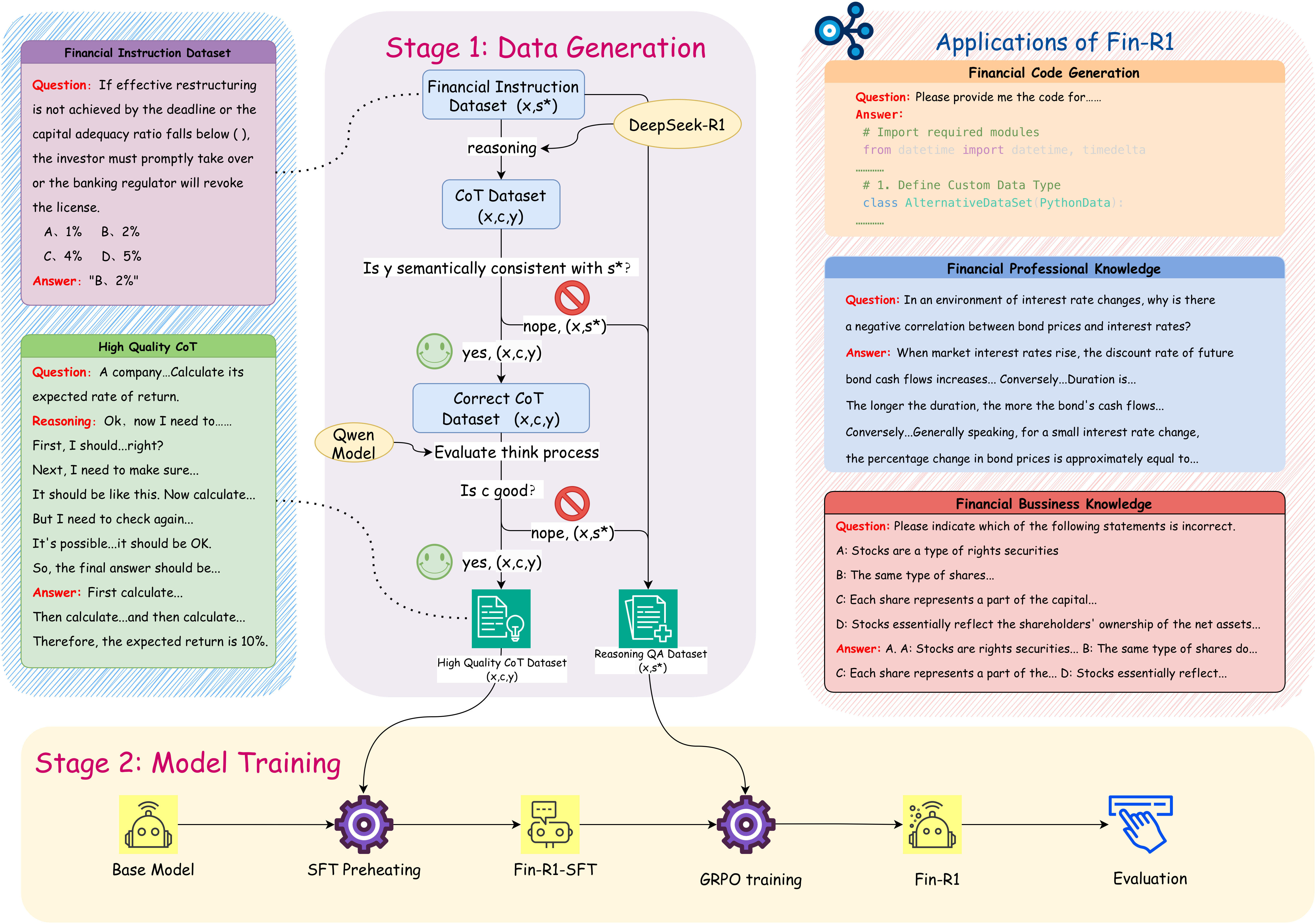}
  \caption{The pipeline for constructing Fin-R1. The diagram depicts the two-stage construction framework of Fin-R1: Data Generation (using DeepSeek-R1 for reasoning to generate CoT data, followed by quality filtering with the Qwen2.5-72B-Instruct) and Model Training (including SFT pretraining and GRPO optimization for Fin-R1). Additionally, the right side highlights the performance of Fin-R1 in financial code generation, professional knowledge, and business knowledge. Here, $x$ denotes the input query, $c$ represents the generated reasoning content, $y$ is the model output answer, and $s^{*}$ corresponds to the ground-truth answer contained in $y$.} 
    \label{fig:2stage} 
\end{sidewaysfigure}
\begin{itemize}[leftmargin=*]
\item In the first stage, we construct a reasoning dataset for finance, termed as \textit{\ul{Fin-R1-Data}}, which covers a wide range of financial scenarios encountered in practice. This stage addresses the first challenge of fragmented data sources and the scarcity of CoT data within financial domains, offering the basis for the subsequent SFT and RL. Specifically, Fin-R1-Data is a high-quality bilingual dataset containing over 60,000 entries, integrating data from diverse sources -- including open-source datasets and proprietary examination problems -- to ensure comprehensive coverage of professional expertise, business practices, and numerical reasoning. Its construction proceeds in two steps. During the \ul{\textit{data distillation}} step, we leverage DeepSeek-R1 \citep{guo2025deepseek} to produce reasoning traces while standardizing answer formats. In the subsequent \ul{\textit{data filtering}} step, we employ Qwen2.5-72B-Instruct \citep{yang2024qwen2} as an evaluator to investigate the logical consistency, coherence, and domain alignment of the data, screen out low-quality samples and retain high-quality data for reasoning. The left panel of Figure~\ref{fig:2stage} depicts a high-quality CoT example after filtering. 

\item In the second stage, we post-train LLMs by applying SFT and RL on Fin-R1-Data to obtain our \ul{\textit{Fin-R1}} model. This stage substantially enhances the pre-trained model's generalization and transferability across diverse business applications, addressing the third challenge. 
Specifically, SFT uses carefully collected CoT samples from Fin-R1-Data to enable the model to ``think before answering'' and to conduct integrated reasoning across legal, economic, and quantitative domains. The next step applies GRPO, a computationally efficient variant of proximal policy optimization \citep[PPO,][]{schulman2017proximal} which evaluates model's outputs using group-relative advantage functions, to ensure both correctness and structural consistency in reasoning chains. Such a two-step post-training approach -- first SFT, then GRPO -- enables a 7B-parameter model to deliver reliable and trustworthy outputs in financial applications. Notably, despite being 100 times smaller than leading frontier reasoning models, Fin-R1 achieved an average score of 75.2 on financial reasoning benchmarks, ranking second overall and 
delivering the nearly best performance on mainstream financial benchmarks \citep[e.g., FinQA,][]{chen-etal-2021-finqaa}. Compared with models of the same 7B scale, Fin-R1 outperformed the existing state-of-the-art models by more than 17 points. 

\item Additionally, the two stages together also tackle the second challenge by making the reasoning process transparent. Specifically, in the data construction stage, we explicitly present reasoning paths in a human-readable format in the training data, to guide the model toward a ``think-then-reason'' paradigm. In the model training stage, we incorporate a format reward function (Equation~\eqref{eq:fmt-reward}) during RL training to further constrain the model’s outputs, ensuring that reasoning remains explicit and interpretable. As an illustration, Figure~\ref{fig:total} presents detailed outputs of Fin-R1 in both Chinese and English, while the right panel of Figure~\ref{fig:2stage} showcases concise examples across various financial scenarios.

\end{itemize}

\textbf{Paper organization}. The remainder of this paper is organized as follows. Section \ref{sec-re} reviews the related work. Section \ref{sec: data-con} presents the construction of Fin-R1-Data, while Section \ref{sec: model-train} details the proposed two-step training pipeline. Section \ref{s3} conducts comprehensive experiments to demonstrate the superior performance of Fin-R1 across multiple financial benchmarks. Section \ref{s4} concludes the study and outlines directions for future research. Additional implementation details, data descriptions, case studies, and experimental settings are provided in the Supplementary Material to facilitate reproducibility.

{\singlespacing	
\section{Related Work}\label{sec-re}
}

In this section, we discuss three strands of related work: pre-training, post-training, and their applications to the financial domain.

\textbf{Pre-training}. Pre-training enables LLMs to learn linguistic patterns and fundamental knowledge from massive text corpora~\citep{devlin2019bert,floridi2020gpt,xue2020mt5}. 
Through auto-regressive next-token prediction on trillions of internet tokens, these models absorbed knowledge across diverse domains such as programming~\citep{kocetkov2022stack}, mathematics~\citep{drori2022neural}, and other professional or scientific fields~\citep{lo2019s2orc}. However, LLM pre-training faces several practical challenges. First, it is highly resource-intensive: for example, training PaLM with 540 billion parameters requires 6,144 TPU v4 chips over several weeks~\citep{chowdhery2023palm}. 
Second, as high-quality internet text data show signs of  exhaustion~\citep{villalobos2022will}, 
the scaling law ~\citep{kaplan2020scaling} that links model size, data volume, and performance appears to approach its limit. 
Although large-scale synthetic corpora can be potentially employed, their effectiveness has not yet been fully validated
~\citep{shumailov2024ai}. 
Third, pre-trained models often struggle to generalize to domain-specific reasoning tasks.  
This 
hinders their abilities to understand complex financial logic, such as derivative pricing or risk-hedging strategies~\citep{zoph2020rethinking,abnar2021exploring}.

\textbf{Post-training}. There is a growing consensus that additional algorithmic paradigms beyond pre-training are required to move closer to AGI. Recent reasoning-oriented models, such as OpenAI’s o1 and DeepSeek-R1, position post-training as such a paradigm~\citep[see][for an overview]{Kumar_2025_LLM}. Post-training aims to mitigate pre-trained models' limitations by aligning their outputs with human values, while reducing biases and inaccuracies~\citep{bai2022constitutional}. It typically includes fine-tuning for task-specific adaptation \citep{trung2024reft}, RL for encouraging the model to explore better outputs guided by feedback signals \citep{Zhang_2025_Survey}, and test-time scaling for boosting model performance without retraining the model~\citep[TTS,][]{hu2022lora}. Among these methods, RL-based post-training has repeatedly demonstrated that well-defined reward signals can drive AI agents to achieve superhuman performance on complex tasks~\citep{ouyang2022training}. 

Among RL algorithms, PPO, a computationally efficient approximation to the trust region policy optimization algorithm~\citep{schulman2015trust}, has been widely adopted for post-training LLMs.
By limiting the divergence between the old and new policies, PPO effectively mitigates the issue of policy collapse commonly encountered in traditional policy gradient algorithms. However, PPO suffers from several limitations, including training instability, high sensitivity to hyperparameter tuning, and substantial computational cost arising from the need to learn a value function for the language model  \citep{engstrom2020implementation, zheng2023secrets, xu2024dpo}.

To address these challenges, \cite{rafailov2023direct} proposed a direct preference optimization (DPO), which expresses the reward model in closed-form using the optimal policy, and transforms the complex RL problem into a more tractable classification task. DPO eliminates the need for value function training, substantially lowers the computational cost, and reduces the number of hyperparameters. Nevertheless, it remains highly dependent on the reference policy and suffers from limited generalization to out-of-distribution samples \citep{xu2024contrastive, feng2024towards, xu2025doubly,ye2025robust}. 

Building on PPO, \citet{shao2024deepseekmath} achieved a major breakthrough in LLM reasoning by introducing the GRPO algorithm.  
Unlike traditional PPO, which primarily relies on a single-stream reward to guide model optimization, GRPO generates multiple candidate outputs and computes advantages based on their relative performance within the group, thus eliminating the need for learning a value network 
and considerably improving the computational efficiency.
Compared to both PPO and DPO, GRPO demonstrates clear advantages 
in reasoning-intensive tasks such as mathematical problem solving \citep{guo2025deepseek}. See \citet{Zhang_2025_Survey} for a comprehensive discussion of RL algorithms for LLMs.

\textbf{Financial LLMs}. 
Pre-trained and post-trained models discussed above are primarily tailored for general-purpose tasks. 
Recently, several financial LLMs have been developed, such as BloombergGPT~\citep{wu2023bloomberggpt}, DISC-FinLLM~\citep{chen2023disc}, PIXIU~\citep{xie2023pixiu}, and XuanYuan~\citep{zhang2023xuanyuan}. 
However, these models are primarily designed for non-reasoning tasks. In contrast, our proposed model, Fin-R1, can reason, which allows it to integrate fragmented financial knowledge across business applications to generate practically useful outputs. 
Meanwhile, models such as XuanYuan-FinX1-Preview~\citep{XuanYuan2024Duxiaoman} and Fino1~\citep{qian2025fino1} adopt o1-like reasoning paradigm to strengthen numerical computation and logical analysis in financial applications. In parallel, FinAgent~\citep{zhang2024multimodal} enhances trading decision making via multimodal tool integration. These efforts collectively push financial LLMs from mere text understanding toward explicit reasoning and decision making. Nevertheless, there remains substantial room to improve reasoning accuracy and transparency, and to enhance model transferability across financial scenarios~\citep{huang2024open,xie2024finben}.

{\singlespacing    
\section{Data Construction}\label{sec: data-con}
}

In this section, we introduce Fin-R1-Data, a high-quality  dataset specifically designed 
for post-training financial LLMs. Section~\ref{sec: data_overview} describes the overall structure of Fin-R1-Data and Section~\ref{sec: data_processing} details the data construction process.

{\singlespacing
\subsection{Data Overview}\label{sec: data_overview}
}

\begin{figure}[t]
    \centering
    \includegraphics[width=0.5\textwidth]{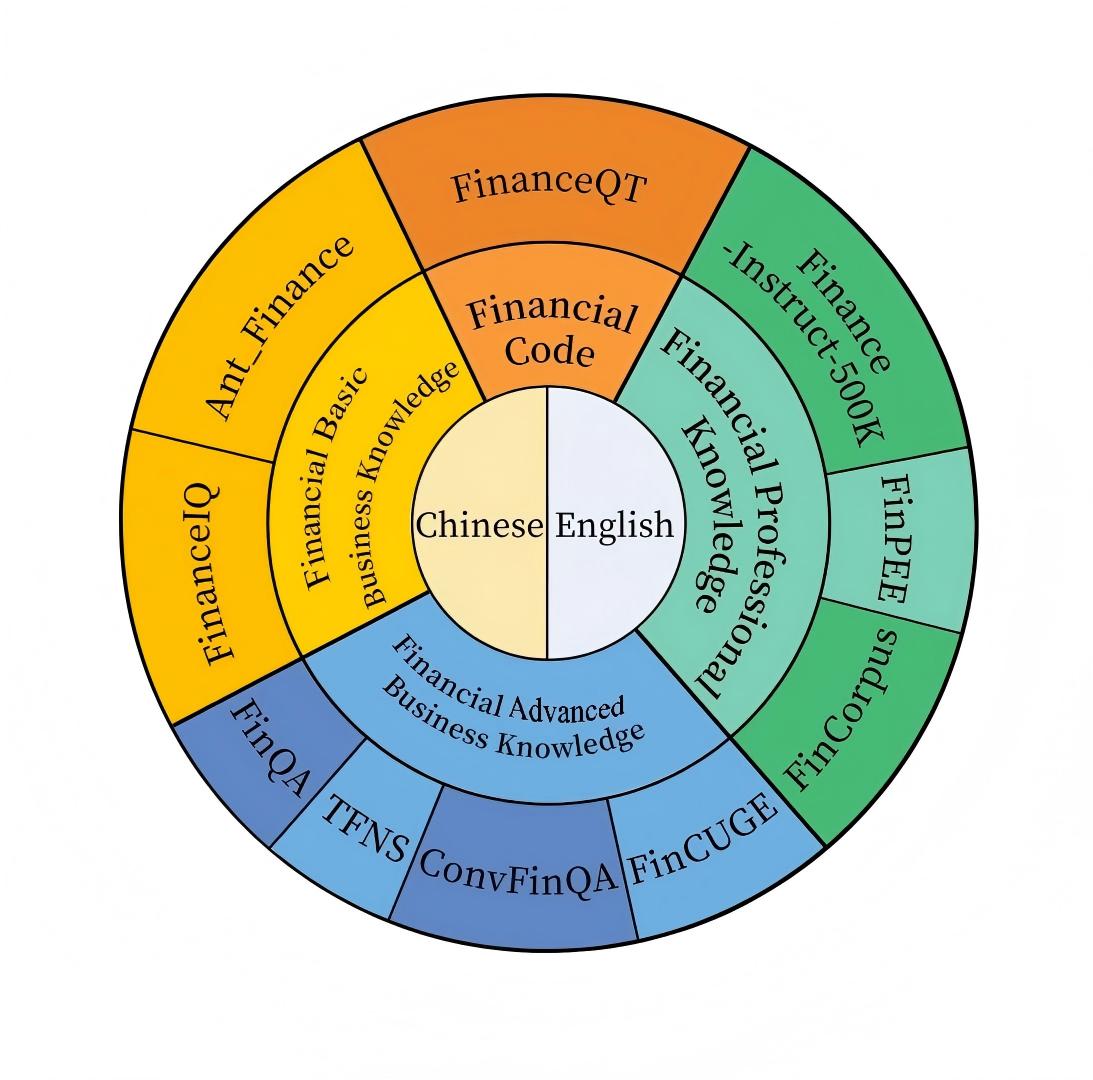}
    \caption{The overall structure of Fin-R1-Data. The inner circle shows the four components of Fin-R1-Data, while the outer circle shows the raw datasets used to construct them.} 
    \label{fig: Data constitution of Fin-R1} 
\end{figure}

Fin-R1-Data consists of 60,091 bilingual (Chinese and English) entries, organized into four categories, as shown in Figure~\ref{fig: Data constitution of Fin-R1}.
We refer to these four categories as \textit{financial advanced business knowledge}, \textit{financial basic business knowledge}, \textit{financial professional knowledge}, and \textit{financial code}, which will be explained in detail as follows.
Each data category is derived by first collecting a raw dataset and then applying specific processing steps. The raw datasets are detailed in Section~\ref{sec:Datasets} of the Supplementary Material. 

\noindent
\textbf{Financial advanced business knowledge}. 
This data subset accounts for roughly $25\%$ of the total data. It primarily consists of problems that require explicit reasoning in financial applications, and often integrates multiple capabilities such as numerical computation, causal inference, and contextual understanding. Some representative examples include numerical reasoning on financial data, sentiment classification of financial news, financial news categorization, and causal relationship extraction.

\noindent
\textbf{Financial basic business knowledge}. 
This data subset primarily focuses on knowledge representation and content generation tasks that do not involve complex reasoning, and contributes over 50\% of the total dataset. It is oriented toward the acquisition, expression, and standardization of basic financial information, covering tasks such as regulatory compliance, financial domain knowledge acquisition, and financial text generation.

\noindent
\textbf{Financial professional knowledge}. This category of data is designed to handle complex, specialized application scenarios, covering questions that can only be answered by humans with sufficient professional financial expertise. It mainly contains  three types of data: (i) explanations of financial terminology, providing precise interpretations of specialized terms and concepts; (ii) financial calculations, involving specialized computational algorithms and models; and (iii) questions from finance-related postgraduate entrance examinations. The last type of data is scarce. To address this, we constructed a financial postgraduate entrance exam (\ul{\textit{FinPEE}}) dataset using original examination questions from Shanghai University of Finance and Economics. This dataset was constructed in two steps: (a) text extraction, where PDF-based source materials were converted into Markdown format using Mineru \citep{wang2024mineru}, and (b) expert review, in which structured question–answer pairs were verified by financial experts to ensure accuracy. 

\noindent
\textbf{Financial code}. The last dataset consists of financial code, i.e., scripts and programs designed to solve financial problems (e.g., quantitative trading). The purpose of constructing this dataset is to enable LLMs to automatically generate financial code and quantitative strategy scripts -- both play a vital role in applications such as algorithmic trading which requires high-frequency code generation and execution, risk modeling which requires coding for building credit risk assessment models, and portfolio management which requires to implement portfolio optimization strategies. 

{\singlespacing
\subsection{Data Processing}\label{sec: data_processing}
}

\begin{sidewaysfigure}[htbp] 
        \centering 
        \includegraphics[width=\textwidth]{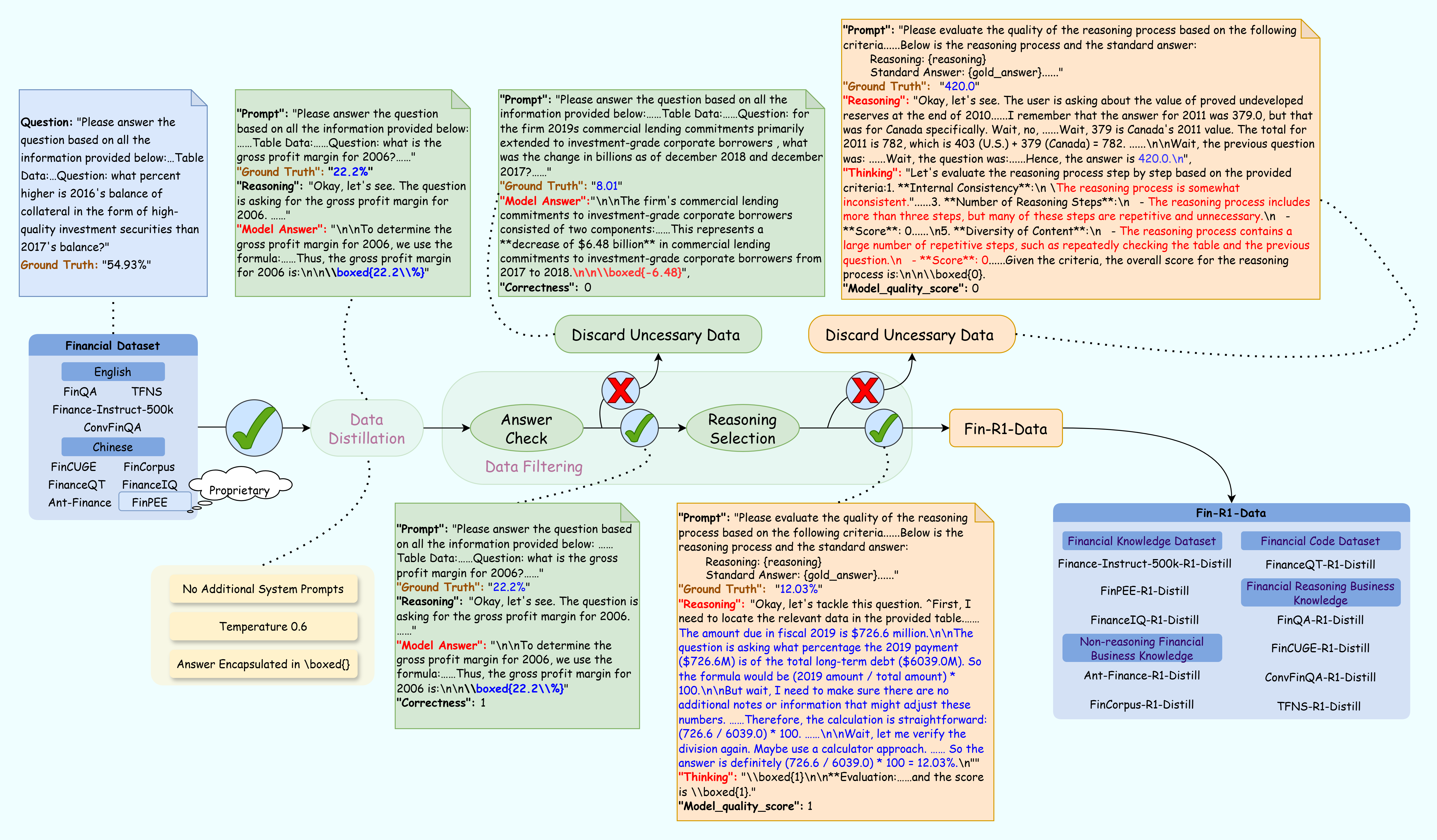}
        \caption{The data construction pipeline. \textbf{Data distillation:} generate the reasoning path and the answer to each question in raw datasets. \textbf{Data filtering:} review the answer and reasoning path. The examples for each step are presented in boxes.}  
        \label{fig: data_construct_pipeline} 
\end{sidewaysfigure}

In this section, we describe how Fin-R1-Data was constructed based on the raw datasets detailed in Section~\ref{sec:Datasets} of the Supplementary Material. Our data construction pipeline consists of two major steps: \ul{\textit{data distillation}} and \ul{\textit{data filtering}}. The first step leverages existing LLMs to generate answers to the questions in the raw datasets, whereas the second step reviews the generated answers to ensure their quality and consistency. Figure~\ref{fig: data_construct_pipeline} visualizes the workflow of these two steps.

\textbf{Data distillation}. 
We extract all questions from the raw datasets and employ DeepSeek-R1-671B to generate both the reasoning paths and corresponding answers for each question. We choose DeepSeek-R1-671B for its large number of parameters and strong reasoning capabilities, which ensure both diversity and reliability in the generated answers.  
When prompting DeepSeek-R1-671B to generate reasoning paths and answers, we configure the hyperparameters in alignment with its official specifications. In particular:
(i) The temperature to generate tokens is set to 0.6. 
(ii) For mathematical questions, we employ the standard prompt ``Please use \textbackslash boxed\{\} to wrap the final answer'' to ensure consistency in the answer format. 
(iii) A line break ``\textbackslash n'' is forced to appended at the beginning of each output. 
Details of the prompts used to generate the answers are provided in Section~\ref{sec: prompt_example_data_distillation} of the Supplementary Material.

\textbf{Data filtering}. 
This step consists of \ul{\textit{answer check}} and \ul{\textit{reasoning selection}}.  
The former evaluates the accuracy of responses generated in the data distillation step, whereas the latter assesses the quality and coherence of the reasoning paths. We illustrate them in Figure~\ref{fig: data_construct_pipeline}.

Specifically, we first review the responses generated by DeepSeek-R1-671B during data distillation. 
For objective questions, we retain only those responses whose answers exactly match the correct ones. For subjective questions, we adopt the LLM-as-Judges paradigm -- using a well-tuned LLM as an automated evaluator -- to access the correctness of each answer and discard any incorrect ones. 
We further conduct a comparative study to compare
GPT-4o \citep{openai2024gpt4o} against Qwen2.5-72B-Instruct \citep{yang2024qwen2} as judges, and evaluate different prompting strategies for enabling these LLMs to serve effectively in this role. Our experimental results show that Qwen2.5-72B-Instruct achieves an accuracy rate of 99.6\% and outperforms GPT-4o. We thus select Qwen2.5-72B-Instruct as the judge model and employ the optimal prompt that achieves the highest accuracy. Further details on the experimental setup and results are provided in Section~\ref{sec: model_selection_answer_check} of the Supplementary Material.

We next focus on the filtered dataset that contains correct answers to the questions and evaluate their reasoning paths. Following~\cite{xie2024finnlpa}, we similarly adopt the LLM-as-Judges paradigm and prompt the LLM to evaluate the reasoning paths according to criteria from the following seven dimensions: internal consistency, term overlap rate, number of reasoning steps, logical coherence, content diversity, task-domain relevance, and alignment with task instructions.
To evaluate the performance of LLMs as judges, we conducted additional experiments to compare human annotations against LLM evaluations. The results, detailed in Section~\ref{sec: model_selection_reasoning_selection} of the Supplementary Material, demonstrate that Qwen2.5-72B-Instruct closely aligns with human judgments, exhibiting only minor discrepancies, whereas GPT-4o shows larger deviations. Based on these findings, we select Qwen2.5-72B-Instruct to assess the quality of the reasoning paths,  
and retain only the high-quality ones for the subsequent SFT. In the yellow boxes of Figure~\ref{fig: data_construct_pipeline}, we present examples of both a high-quality and a low-quality reasoning trajectory to illustrate their differences. The sentence beginning with ``Thinking:'' is generated by Qwen2.5-72B-Instruct, which analyzes the rationality of each reasoning trajectory and assigns a corresponding quality score.

\section{Model Training}\label{sec: model-train}

\begin{figure}[t] 
        \centering 
        \includegraphics[width=0.8\textwidth]{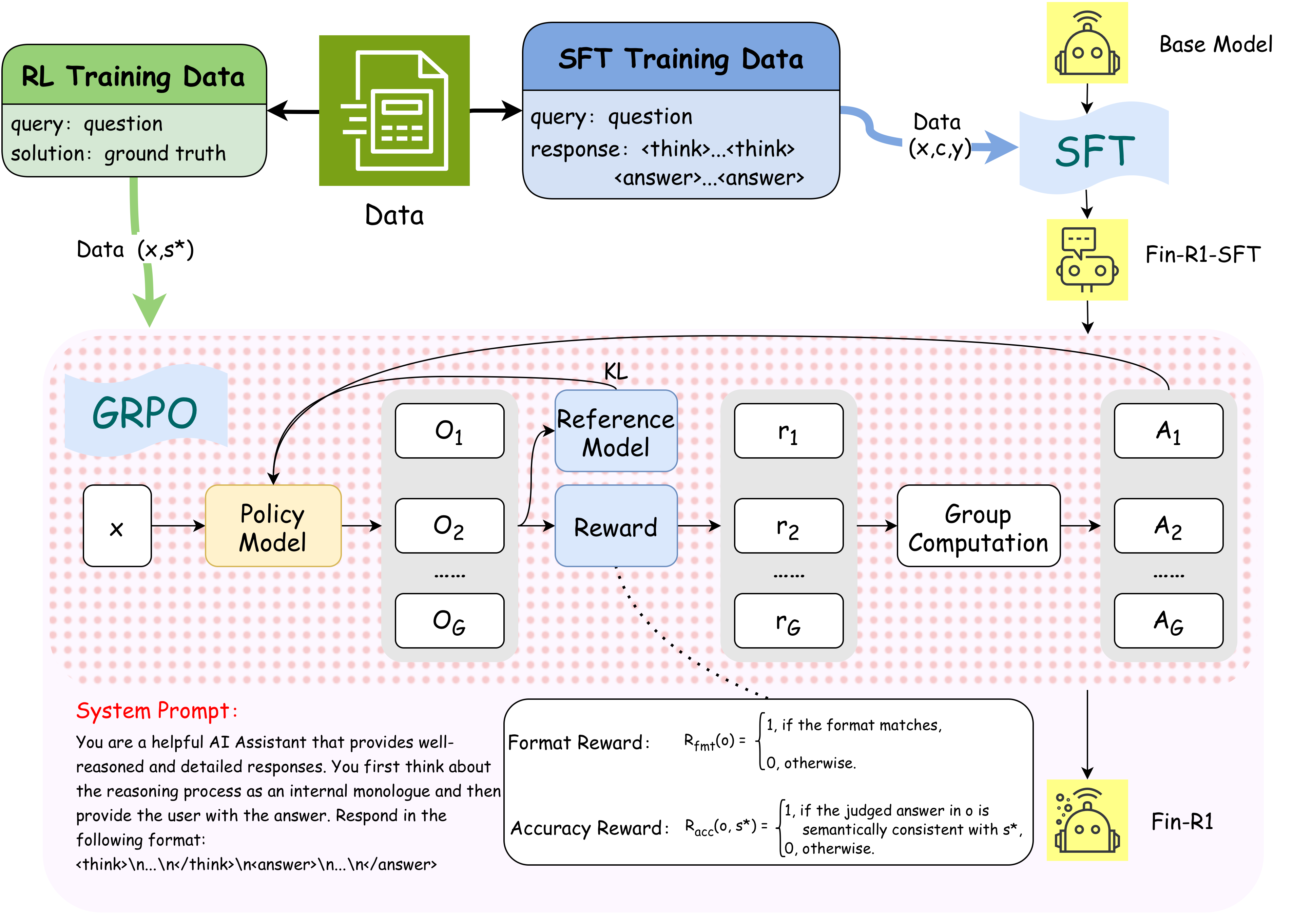}
        \caption{Stage 2 -- The post-training pipeline. The model is first trained via SFT on Fin-R1-Data and is then optimized through GRPO with two reward functions, a format reward and an accuracy reward.} 
        \label{train} 
\end{figure}

In this section, we detail our training procedure for Fin-R1. We adopt Qwen2.5-7B-Instruct as the base model. 
To enhance its reasoning capabilities in finance, we adopt a two-step post-training procedure, consisting of SFT and RL. In the first step, the model is trained on the Fin-R1-Data to learn to think before answering and to expand its financial-domain knowledge. Building on this, GRPO is employed in the second step with two rule-based reward functions -- a \textit{format reward} and an  \textit{accuracy reward} -- to further enhance the model’s reasoning capabilities. Figure~\ref{train} visually summarizes this training procedure. 

\subsection{Supervised Fine-tuning}
As an autoregressive model, an LLM generates text by sequentially predicting the next token conditioned on the previously generated token sequence. Each token $v$ corresponds to a word, subword, or punctuation mark drawn from a  vocabulary $\mathcal{V}$. Let $\pi_\theta$ denotes the probability mass function of the model's next-token distribution, i.e., 
\begin{equation*}
    \pi_{\theta}(v|v_{-})=\mathbb{P}(\textrm{LLM~selects~}v\textrm{~as~the~next~token}|\textrm{the~previous~token~sequence}~v_{-}),
\end{equation*}
where $\theta$ denotes the model parameter. We refer $\pi_{\theta}$ as the policy as it is  the objective being optimized during the RL training.

Pre-trained autoregressive models are often limited, as they tend to mimic their pre-training data (e.g., by repeating the query), which reduces their effectiveness in various  downstream tasks. To address this, SFT is employed to inject desired behavior into the model. Its core idea is to leverage the model's existing ability to generate coherent text and adapt it to a new task by fine-tuning $\theta$ with high-quality supervised examples, through which the model is explicitly guided to output the desired solutions. This approach avoids training a model from scratch while enhancing its performance on the target task. 

\begin{figure}[t] 
        \centering 
        \includegraphics[width=0.7\textwidth]{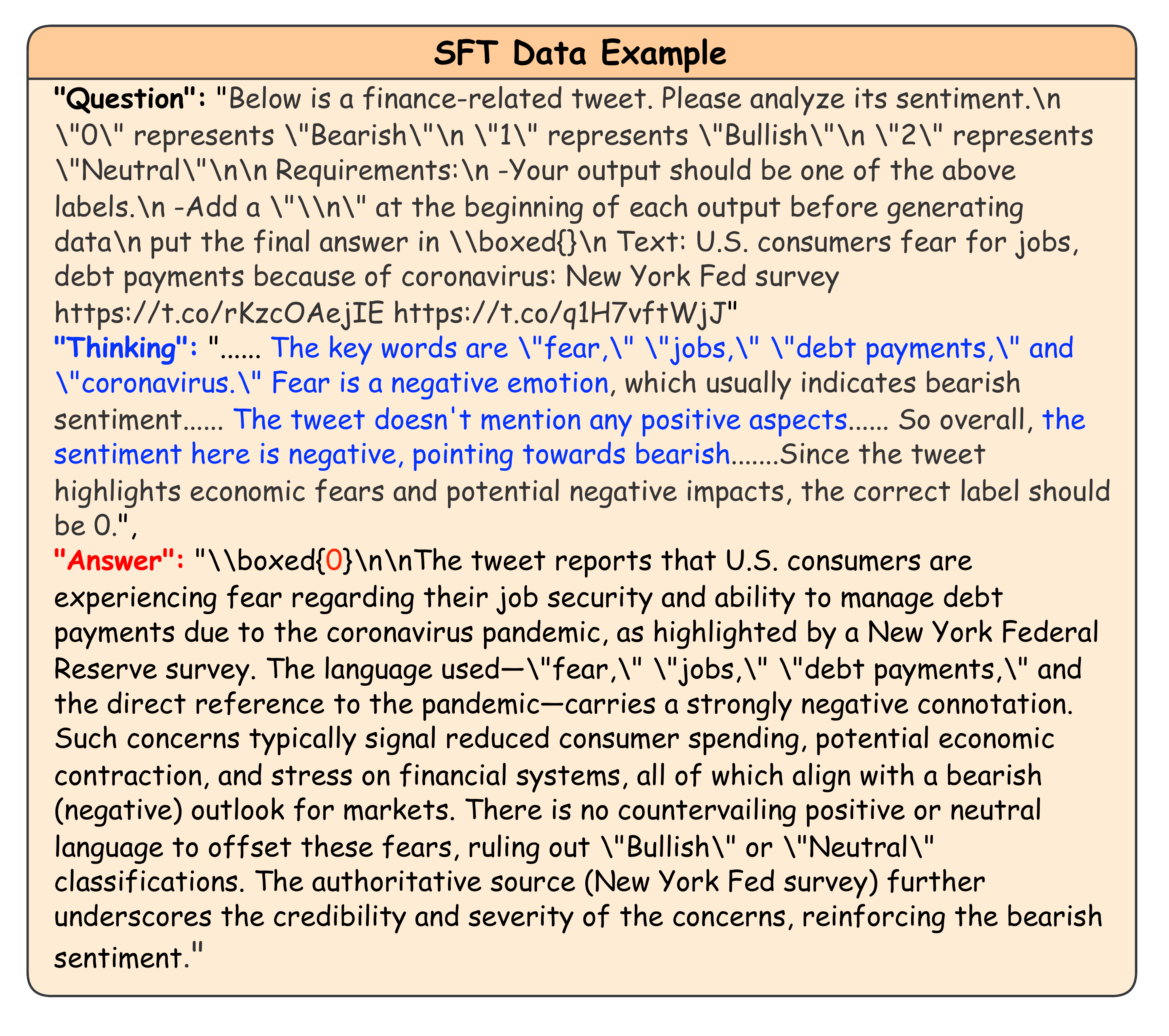}
        \caption{An SFT data example. The input $x$ is the question, and the model output $o$ includes a reasoning trace $c$ (enclosed in the Thinking block) and an answer $y$ (enclosed in the Answer block). The answer provides a numerical solution $s$, which in this case is 0.}
        \label{fig: sft-temp} 
\end{figure}

More specifically, we denote the SFT training dataset by 
\(\mathcal{D}_{\mathrm{SFT}}\), consisting of pairs $(x,o)$, where $x$ is a query and $o$ is the corresponding output. Each output $o$ consists of two parts: a reasoning trace \(c\), 
enclosed in \texttt{<think>...</think>} tags, and an answer \(y\), enclosed in \texttt{<answer>...</answer>} tags. 
It is worth noting that, for objective questions, the answer typically includes a numerical solution (denoted by $s$), often highlighted using \texttt{\textbackslash boxed{...}}, along with any supporting explanation. Figure~\ref{fig: sft-temp} illustrates an example of the data structure.

SFT's objective function is typically set to the cross-entropy loss that minimizes the discrepancy between the model’s predicted next token and the corresponding ground-truth token in the SFT dataset. 
We decompose a response $o$ into a sequence of tokens \((o_1,\ldots,o_n)\). The probability of outputting \(o\) 
given the query \(x\) can be factorizes as
\[
\pi_\theta(o \mid x)
=
\prod_{t=1}^{n}
\pi_\theta\!\left(o_t \mid x, o_{<t}\right),
\]
where $o_{<t}:=\{o_1,\ldots,o_{t-1}\}$ for $t\leq n$.
To compute $\theta$, SFT minimizes the following cross-entropy loss, which is equivalent to maximizing the log-likelihood: 
\begin{equation}\label{eq:sft-loss}
\mathcal{L}_{\mathrm{SFT}}(\theta)
~=~
\sum_{(x,o)\sim \mathcal{D}_{\mathrm{SFT}}}
\bigl[-\log \pi_\theta(o\mid x)\bigr].
\end{equation}
By definition, the use of the cross-entropy loss function \eqref{eq:sft-loss} encourages the model to assign higher probability to the high-quality answers provided in the dataset. In doing so, it guides the model to internalize correct reasoning and answer patterns, effectively adapting it to the target domain task. 
In practice, the minimization of \eqref{eq:sft-loss} is carried out via 
stochastic gradient optimization where gradients are obtained through 
backpropagation, and the parameters are updated using adaptive algorithms such as 
Adam \citep{kingma2014adam}.

To conclude this section, we remark that SFT is generally understood as a method of behavior cloning: it aims to imitate expert demonstrations -- including both the reasoning traces and the final answers -- but it cannot learn entirely new behaviors beyond those present in the training data. This is its fundamental limitation. In addition, heterogeneity in annotators’ writing styles and preferences may further reduce its effectiveness. These limitations highlight the need to move beyond behavior cloning and enable the model to refine its behavior through feedback.

\medskip

\subsection{Group Relative Policy Optimization}
We employ RL to address the aforementioned limitations of SFT. The empirical benefits of incorporating RL are illustrated in Table~\ref{tab:model_ablation}, which shows a substantial improvement in model performance. 
Among RL algorithms for LLM reasoning, GRPO is particularly attractive because it optimizes the policy using relative rewards within a group and 
avoids the need to compute value networks as in PPO-type algorithms, which makes it more computationally efficient while maintaining statistical efficiency comparable to PPO. 

\begin{figure}[t] 
        \centering 
        \includegraphics[width=0.8\textwidth]{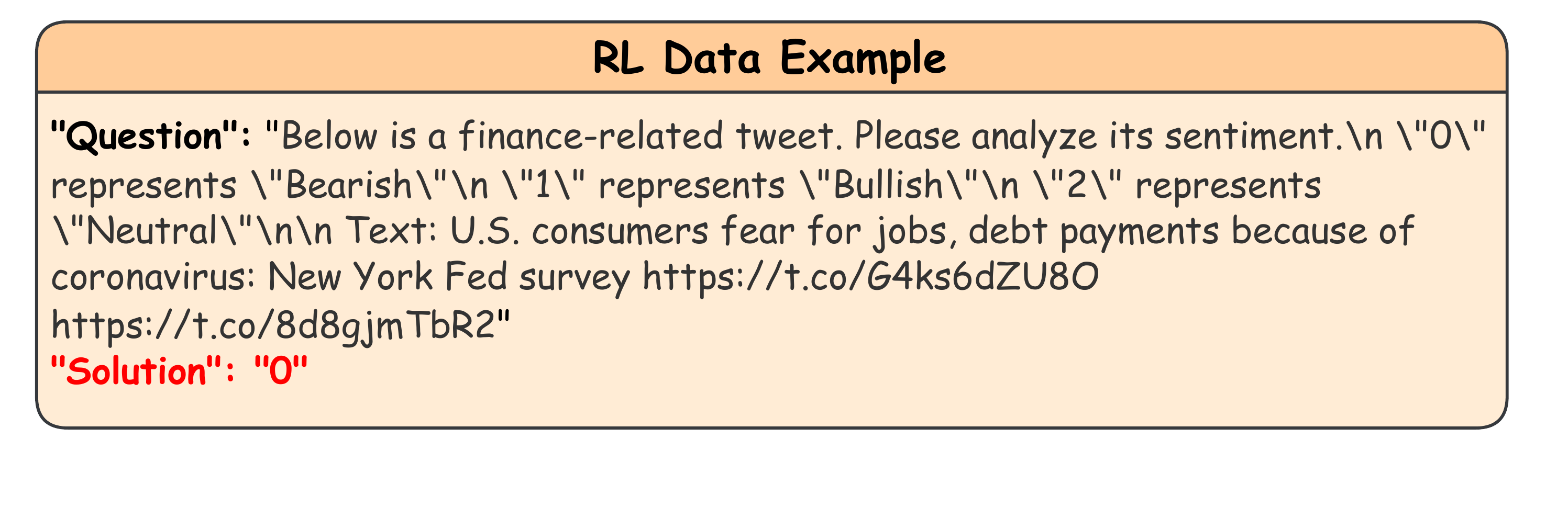}
        \caption{An RL data example, consisting of an objective question $x$ and its numerical solution $s^*$.} 
        \label{fig: rl-temp} 
\end{figure}

We begin by introducing the notation used throughout this section.  
The RL training dataset 
$\mathcal{D}_{\mathrm{RL}}$ consists of pairs  $\{(x, s^{*})\}$,
where $x$ denotes an objective question, and $s^*$ denotes its numerical solution; see Figure \ref{fig: rl-temp} for a concrete data example. We highlight three differences between this dataset and the SFT training data: (i) In SFT, the query $x$ includes both objective and subjective questions, whereas in the RL dataset, $x$ is restricted to objective questions only; (ii) In SFT, each data entry contains the full output $o$, including the reasoning trace $c$, as well as the answer $y$, which comprises both the solution $s^*$ and its explanation. In contrast, the RL data contains only the solution $s^*$; (iii) Because SFT requires the full output $o$, we apply data filtering (see Section \ref{sec: data_processing}) to remove instances in which Qwen's generated output $o$ is inconsistent with $s^{*}$. As a result, the SFT dataset does not contain these hard questions with incorrect outputs, whereas GRPO uses these  questions for training.

Given the dataset, GRPO operates as follows: it initialize the policy model $\pi_{\theta}$ to the SFT policy $\pi_{\textrm{SFT}}$, iteratively generates multiple candidate outputs $o$ for each question $x$, evaluates these outputs against the solution $s^{*}$ using predefined reward functions, computes the advantage function from these rewards, and updates the policy accordingly. See Algorithm \ref{alg:grpo} for its pseudocode and Figure \ref{train} for a visualization. We detail the output rewards, advantage calculation, and policy update below.

\begin{algorithm}[t]
\small
\caption{Group Relative Policy Optimization}\label{alg:grpo}
\textbf{Input:} 
SFT policy $\pi_{\mathrm{SFT}}$; 
RL dataset $\mathcal{D}_{\mathrm{RL}}$; 
hyperparameters $(I, M, G, \epsilon, \beta)$

\begin{algorithmic}[1]
\State Initialize policy $\pi_{\theta} \leftarrow \pi_{{\mathrm{SFT}}}$

\For{iteration $=1,\dots,I$}  

    \For{step $=1,\dots,M$}
        \State Sample a minibatch $\mathcal{B}$ from $\mathcal{D}_{\mathrm{RL}}$
        \State Freeze the old policy $\pi_{\theta_{\mathrm{old}}} \leftarrow \pi_{\theta}$

        \For{each $(x,s^*) \in \mathcal{B}$}
            \State Sample $G$ candidate outputs
            \[
                \{o_i\}_{i=1}^{G} \sim \pi_{\theta_{\mathrm{old}}}(\cdot \mid x)
            \]
            \State Compute rewards $r_i = R(o_i, y^*_i)$ using format and accuracy criteria
            \State Estimate group-relative advantages $\widehat{A}_{i,t}$
        \EndFor

        \State Update the policy parameter $\theta$ via gradient ascent with the gradient in Equation~\eqref{eq:grpo}
    \EndFor
\EndFor

\State \textbf{return} final policy $\pi_{\theta_{\text{RL}}}$
\end{algorithmic}
\end{algorithm}

\textbf{Reward function}. 
Given a training data entry \((x, s^*)\) sampled from 
\(\mathcal{D}_{\mathrm{RL}}\), GRPO uses its sampling policy (denoted by $\pi_{\theta_{\mathrm{old}}}$) to generate 
$G\ge 2$ candidate outputs $\{o_1, \ldots, o_G\} \sim \pi_{\theta_{\mathrm{old}}}(\cdot \mid x)$ to the question \(x\).   
For each output $o_i$, its reward $r_i$ is given by
\[
r_i = R(o_i,s^*)
    = R_{\mathrm{fmt}}(o_i)
    + R_{\mathrm{acc}}(o_i,\, s^*),
\]
where \(R_{\mathrm{fmt}}\) (defined in \eqref{eq:fmt-reward}) denotes the format reward function that enforces strict 
constraints on the format of the output, and \(R_{\mathrm{acc}}\) (defined in \eqref{eq:acc-reward}) denotes the accuracy reward function that evaluates semantic agreement between \(o_i\) and the ground-truth solution 
\(s^*\).

The format reward encourages outputs outputs to include a reasoning trace enclosed within \texttt{<think>}$\ldots$\texttt{</think>} tags and a concise answer enclosed within \texttt{<answer>}$\ldots$\texttt{</answer>} tags, without additional content outside these tags. It assigns a score of 1 if the output strictly follows this format and 0 otherwise, i.e., 
\begin{equation}\label{eq:fmt-reward}
   R_{\text{fmt}}(o) = 
\begin{cases}
1, & \text{if the format matches,} \\
0, & \text{otherwise.}
\end{cases} 
\end{equation}

To compute the accuracy reward, we employ Qwen2.5-Max \citep{qwen2024} as the evaluator. Specifically, for each candidate output $o$, we extract the answer enclosed within the \texttt{<answer>}$\ldots$\texttt{</answer>} tags and prompt Qwen2.5-Max to assess whether the enclosed answer is semantically consistent with the ground-truth answer $s^{*}$ (See Section~\ref{sec:Prompt of Judging} of the Supplementary Material for the prompts used to instruct Qwen2.5-Max as the evaluator). If they are consistent, the reward is set to $1$; otherwise, it is set to $0$. Formally, we define
\begin{equation}\label{eq:acc-reward}
R_{\mathrm{acc}}(o, s^*) =
\begin{cases}
1, & \text{if the judged answer in } o \text{ is semantically consistent with } s^*,\\[2pt]
0, & \text{otherwise.}
\end{cases}
\end{equation}

\textbf{Advantage Calculation}. Based on the reward $r_i$, PPO estimates the value function
$V^{\pi_{\theta}}(x) = \mathbb{E}(r_i \mid x)$,
where the expectation is taken over the stochasticity in generating the output  induced by the policy $\pi_{\theta}$, which is typically random. It then calculates the advantage function as the difference between $r_i$ and the estimated value to construct the gradient to update $\theta$. The purpose of using the advantage rather than the raw reward is to reduce the variance of the gradient estimate, which in turn improves the statistical efficiency of policy learning  \citep{williams1992simple}. However, the value function is typically parameterized by a transformer network, whose training and storage incur substantial computational and memory overhead. 

GRPO addresses this challenge by completely avoiding value function estimation. Instead, for each input $x$, it generates $G$ rewards to approximate the value using their empirical mean, $\bar{r} = G^{-1}\sum_{i=1}^{G} r_i$.
More specifically, their advantage function for each $r_i$ is calculated as follows:
\begin{equation}\label{eq:advance-func}
    A_i = \frac{r_i - \bar{r}}{\text{max}(\epsilon,\sqrt{\frac{1}{G} \sum^G_{j=1}(r_j - \bar{r})^2})},
\end{equation}
where $\epsilon$ is a small constant (e.g., $10^{-8}$) used to prevent division by zero. As shown in \eqref{eq:advance-func}, in addition to computing the difference between $r_i$ and its expected value, the advantage is further standardized using the standard error of $\bar{r}$. This standardization increases the weight assigned to differences with low uncertainty in $\bar{r}$ while downweighting those associated with higher uncertainty. For sufficiently large $G$, $\bar{r}$ approaches the oracle value function, making GRPO as statistically efficient as PPO while being considerably more computationally efficient.

\textbf{Policy Update}. At each training step, GRPO samples a minibatch $\mathcal{B}$ of $(x, s^{*})$ pairs from $\mathcal{D}_{\mathrm{RL}}$,  generates $G$ outputs for each $x\in \mathcal{B}$ and computes the advantage $A_i$ for each of the $i$th output $o_i$. It then  calculates the gradient of the policy value $\mathbb{E}_{(x,s^*)\sim \mathcal{D}_{\mathrm{RL}},o\sim \pi_{\theta}(\bullet|x)} R(o,s^*)$
as follows,
\begin{equation}\label{eq:grpo}
\begin{aligned}
    g(\theta)=\frac{1}{G} \sum_{i=1}^G  \frac{1}{|o_i|}\sum_{t=1}^{|o_i|} \Big\{ \min\big[ w_{i,t}(\theta)  A_{i}, \, &\text{clip}\big( w_{i,t}(\theta), 1 - \epsilon, 1 + \epsilon \big) A_{i} \big]\nabla_{\theta}\log \pi_{\theta}(o_{i,t}|x,o_{i,<t})\\ 
    & - \beta \nabla_{\theta} D_{\text{KL}}\!\left(\pi_{\theta} \,\|\, \pi_{\text{ref}}\right) \Big\}.
\end{aligned}
\end{equation}
Let us elaborate on Equation \eqref{eq:grpo} below. First, $w_{i,t}(\theta)$ denotes the importance sampling (IS) ratio 
\[
w_{i,t}(\theta)=\frac{\pi_{\theta}(o_{i,t}|x,o_{i,<t})}{\pi_{{\theta_{{\mathrm{old}}}}}(o_{i,t}|x,o_{{i,<t}})},
\]
which corrects for the distribution shift between the current policy $\pi_{\theta}$ and the sampling policy $\pi_{\theta_{\textrm{old}}}$. This allows the policy gradient to be evaluated for 
$\theta\neq \theta_{\textrm{old}}$, even though the advantages $A_i$ are computed using samples generated by $\pi_{\theta_\textrm{old}}$.

Second, the clipping operator \(\text{clip}\!\left( w_{i,t}(\theta), 1 - \epsilon, 1 + \epsilon \right)\) restricts the IS ratio to the range \([1 - \epsilon, 1 + \epsilon]\), which prevents avoid large ratios that would inflate the variance of the policy gradient.

Finally, $\nabla_{\theta} D_{\text{KL}}\!\left(\pi_{\theta} \,\|\, \pi_{\text{SFT}}\right)$ denotes the gradient of the Kullback–Leibler (KL) divergence measure between $\pi_{\theta}$ and the SFT policy,  estimated using the minibatch of samples, and $\beta>0$ denotes the regularization parameter. This KL regularization term encourages the learned policy to remain close to the SFT policy, limits excessive exploration, preserves knowledge acquired during SFT, and mitigates the risk of training collapse. 

Given the gradient in \eqref{eq:grpo}, GRPO performs the gradient ascent algorithm $\theta\leftarrow \theta+\eta g(\theta)$ $M\ge 1$ steps, initialized from  $\theta=\theta_{\textrm{old}}$. After these updates, the sampling parameter is updated by setting $\theta_{\textrm{old}}\leftarrow \theta$, and the procedure is repeated.

\section{Experiment}\label{s3}
In this section, we conduct numerical experiments to demonstrate that Fin-R1 -- despite having only 7B parameters -- achieves strong performance across mainstream financial benchmarks. We first introduce the datasets used for evaluation, along with the evaluation methodology. We next describe the baseline models for comparison. Finally, we report our experimental results.

\subsection{Evaluation Datasets and Methodology}

We employ five representative open-source datasets for evaluation: FinQA, ConvFinQA, Ant-Finance, TFNS, and Finance-Instruct-500k. These datasets were chosen for their diversity to ensure a broad and comprehensive evaluation. Except for Finance-Instruct-500k, other datasets are composed of objective questions with unique reference answers. Further details about these datasets can be found in Section~\ref{sec:Datasets} of the Supplementary Material. For Finance-Instruct-500k, we evaluate the performance various models on a custom 10\% test subset, extracted via stratified sampling from the full dataset. For all other datasets, we randomly sample 1,000 questions for testing; if a set contains fewer than 1,000 questions, we employ all questions for testing.

\begin{figure}[t]
    \centering
    \begin{minipage}{0.45\textwidth}
        \centering
        \includegraphics[width=\textwidth]{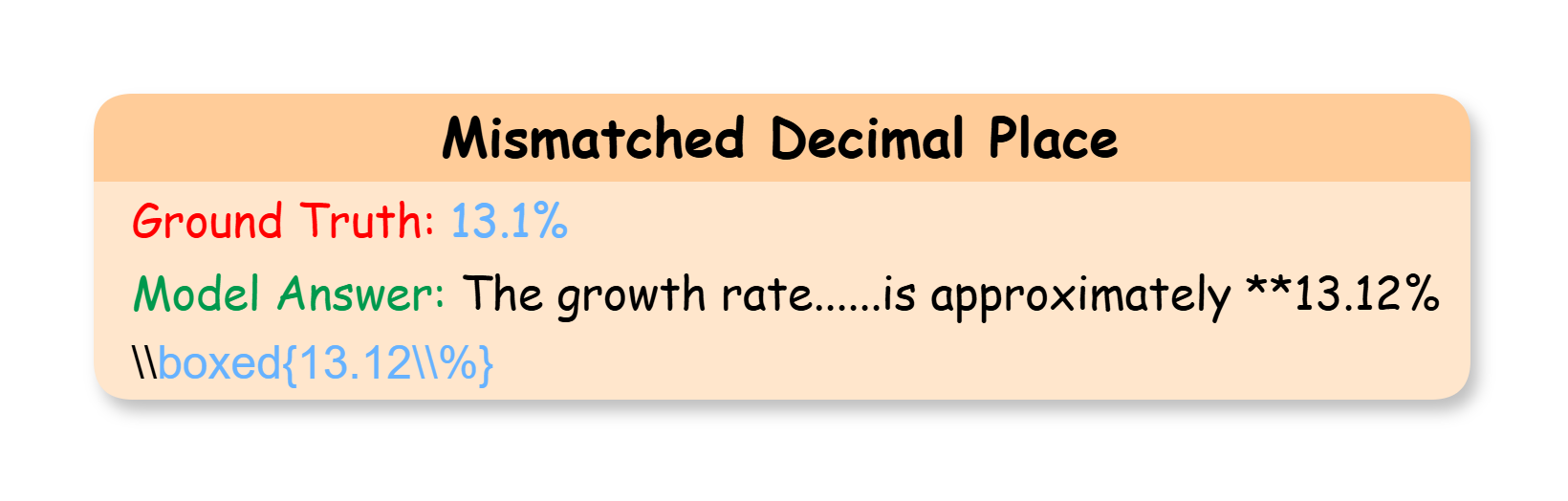}
        \subcaption{Difference in decimal places.}
        \label{decimal}
    \end{minipage}
    \hfill
    \begin{minipage}{0.45\textwidth}
        \centering
        \includegraphics[width=\textwidth]{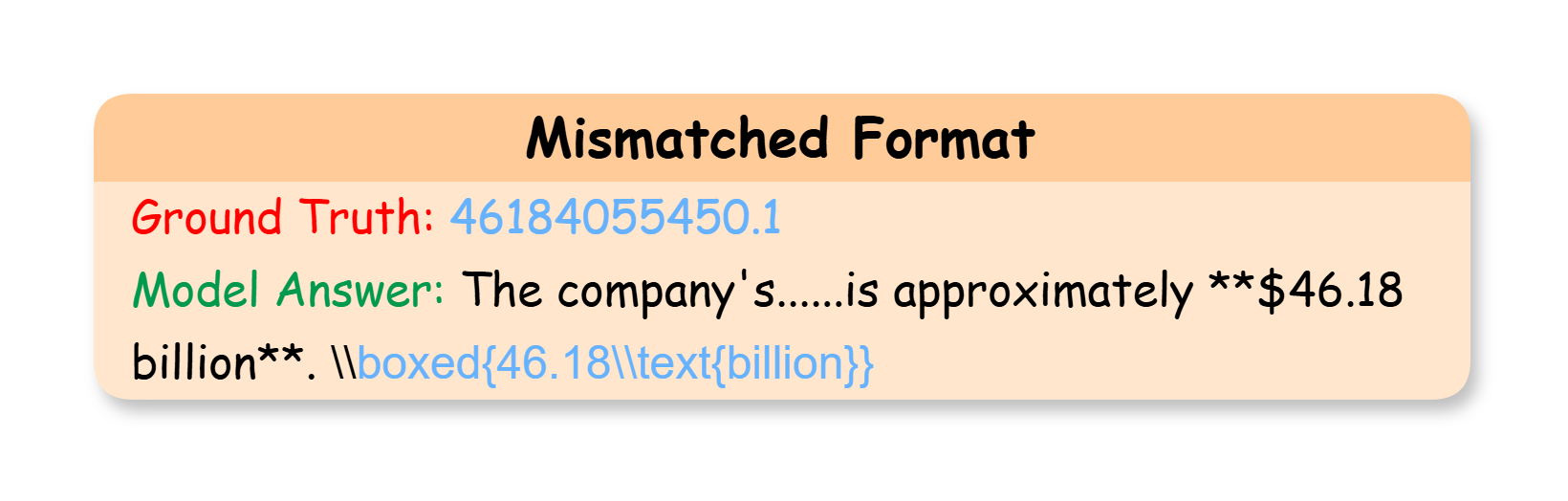}
        \subcaption{Difference in expression.}
        \label{unit}
    \end{minipage}
    \caption{The difference between the model output and the reference solution. Figure \ref{decimal} illustrates the difference in decimal precision,  whereas Figure \ref{unit} visualizes the difference caused by alternative numeric representations.}
\label{badcases}
\end{figure}

For numerical calculation questions, LLM-generated answers may be mathematically correct but expressed in a different valid format, due to variations in decimal precision or alternative numeric representations (see Figure \ref{badcases} for illustrations). Directly comparing these answers with the reference answers will often mark them as incorrect because they do not match exactly, even though they should be considered correct. To address this, we use LLMs as an automated evaluation judge for answer check by adopting the prompt template and the evaluation methodology proposed by \cite{zhu2024judgelm}. 
Although this evaluation paradigm 
appears straightforward, we have systematically conducted numerical experiments to fine-tune several critical parameters in the prompt template for optimizing the reliability of the LLM-based judge. 
Further experimental details can be found in the Section \ref{sec:Prompt of Judging}  of the Supplementary Material. Based on the optimized prompt, we instruct the LLM to report each model’s performance on a 100-point scale, which reflects the percentage of questions answered correctly by each model.

\subsection{Baselines}
We compare our Fin-R1 against \textbf{eight} state-of-the-art LLMs, including (i) DeepSeek-R1; (ii) DeepSeek-R1-Distill-Qwen-7B; (iii) DeepSeek-R1-Distill-Qwen-14B; (iv) DeepSeek-R1-Distill-Qwen-32B; (v) DeepSeek-R1-Distill-Llama-70B; (vi) Qwen-2.5-7B-Instruct; (vii) Qwen-2.5-14B-Instruct; and (viii) Qwen-2.5-32B-Instruct. The models labeled ``Distill'' are lightweight variants obtained through knowledge distillation, a technique that transfers knowledge from a large, high-performance teacher model (e.g., DeepSeek-R1) to smaller student models (e.g., Qwen and Llama series), while preserving comparable reasoning capabilities~\citep[see, e.g., Section~2.4 in][]{guo2025deepseek}. This approach enables the efficient deployment of advanced financial LLMs in resource-constrained settings. These baseline models were chosen to cover a broad range of sizes, from lightweight to large-scale architectures, while maintaining strong performance in terms of reasoning capability and computational efficiency. Our comparison aims to comprehensively evaluate Fin-R1 against leading baselines in financial applications.

\subsection{Results}
\begin{table*}[t]
\caption{Evaluation results across different financial benchmarks.}
\label{tab:model_compare}
\vspace{10pt}
\resizebox{1.0\textwidth}{!}{
    \centering
    \renewcommand{\arraystretch}{1.5} 
    \begin{tabular}{>{\Large}l >{\Large}c *{5}{>{\Large}c} >{\Large}c} 
    \toprule[2pt]
    \textbf{Model} & \textbf{Parameters} & \textbf{FinQA} & \textbf{ConvFinQA} & \textbf{Ant\_Finance} & \textbf{TFNS} & \textbf{Finance-Instruct-500K} & \textbf{Average} \\
    \midrule[1pt]
    DeepSeek-R1 & 671B & 71.0 & 82.0 & \textbf{90.0} & 78.0 & \textbf{70.0} & \textbf{78.2} \\
    Qwen2.5-32B-Instruct & 32B & 72.0 & 78.0 & 84.0 & 77.0 & 58.0 & 73.8 \\
    DeepSeek-R1-Distill-Qwen-32B & 32B & 70.0 & 72.0 & 87.0 & \textbf{79.0} & 54.0 & 72.4 \\
    Qwen2.5-14B-Instruct & 14B & 68.0 & 77.0 & 84.0 & 72.0 & 56.0 & 71.4 \\
    DeepSeek-R1-Distill-Llama-70B & 70B & 68.0 & 74.0 & 84.0 & 62.0 & 56.0 & 69.2 \\
    DeepSeek-R1-Distill-Qwen-14B & 14B & 62.0 & 73.0 & 82.0 & 65.0 & 49.0 & 66.2 \\
    Qwen2.5-7B-Instruct & 7B & 60.0 & 66.0 & 85.0 & 68.0 & 49.0 & 65.6 \\
    DeepSeek-R1-Distill-Qwen-7B & 7B & 55.0 & 62.0 & 71.0 & 60.0 & 42.0 & 58.0 \\
    \midrule[1pt]
    \textbf{Fin-R1} & \textbf{7B} & \textbf{76.0} & \textbf{85.0} & 81.0 & 71.0 & 62.9 & 75.2 \\
    
    \bottomrule[2pt]
    \end{tabular}
}
\centering
\end{table*}
The results are reported in Table \ref{tab:model_compare}. It can be seen that Fin-R1 delivers impressive performance despite its compact 7B parameter scale. Specifically, it achieves an average score of 75.2, ranking the second overall, and outperforms all models of similar size, with only a 3-point gap from the best-performing DeepSeek-R1 (78.2). Notably, it outperforms DeepSeek-R1-Distill-Llama-70B (69.2) by 6 points. Fin-R1 also attains the first place in two reasoning tasks -- FinQA and ConvFinQA -- with scores of 76.0 and 85.0, respectively, outperforming all competing models. These results demonstrate Fin-R1’s strong reasoning capabilities in financial applications. For the remaining tasks, it also achieves comparable or better performance over baseline models such as Qwen2.5-7B-Instruct.

\begin{figure}[t]
    \centering 
    
    \begin{subfigure}[b]{\textwidth} 
        \centering
        \includegraphics[width=0.7\textwidth]{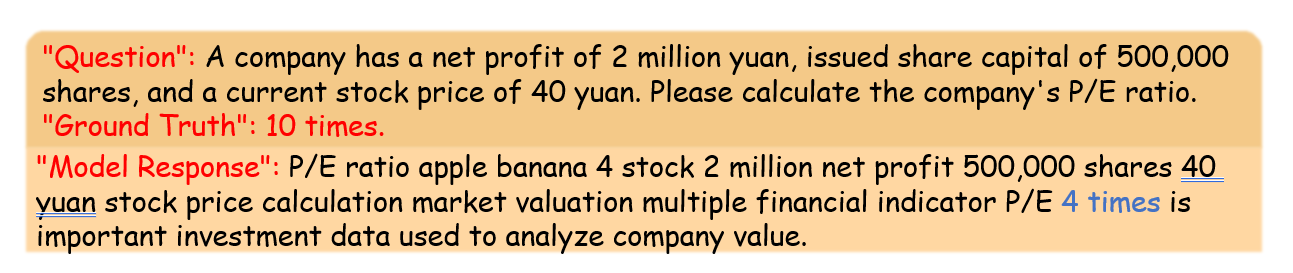} 
        \caption{Fin-R1-Zero output}
        \label{ablationstudy1}
    \end{subfigure}
    
    \vspace{10pt} 

    \begin{subfigure}[b]{\textwidth}
        \centering
        \includegraphics[width=0.7\textwidth]{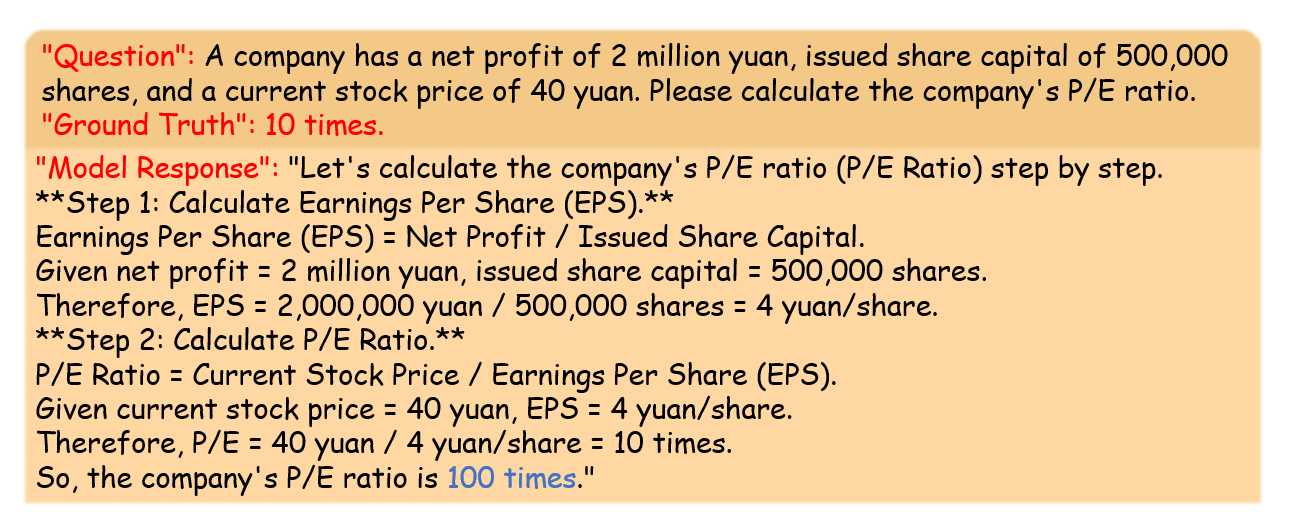}
        \caption{Fin-R1-SFT output}
        \label{ablationstudy2}
    \end{subfigure}
    
    \caption{Example model outputs from the ablation study: Figure~\ref{ablationstudy1} and Figure~\ref{ablationstudy2} show the outputs of Fin-R1-Zero, trained solely on the GRPO algorithm, and Fin-R1-SFT, trained solely on SFT, respectively. The incorrect answers are marked in blue.}
    \label{ablationstudy}
\end{figure}

Next, we conduct an ablation study to demonstrate the effectiveness of our two-step training process (i.e., SFT \& GRPO). %
\begin{table*}[t]
\caption{Evaluation results in different financial benchmarks.}
\label{tab:model_ablation}
\vspace{10pt}
\resizebox{1.0\textwidth}{!}{
    \centering
    \renewcommand{\arraystretch}{1.5} 
    \begin{tabular}{>{\Large}l >{\Large}c *{5}{>{\Large}c} >{\Large}c} 
    \toprule[2pt]
    \textbf{Model} & \textbf{Parameters} & \textbf{FinQA} & \textbf{ConvFinQA} & \textbf{Ant\_Finance} & \textbf{TFNS} & \textbf{Finance-Instruct-500K} & \textbf{Average} \\
    \midrule[1pt]
    Qwen2.5-7B-Instruct & 7B & 60.0 & 66.0 & \textbf{85.0} & 68.0 & 49.0 & 65.6 \\
    Fin-R1-Zero  & 7B & 62.1  & 76.3 & 72.2 &  70.9  &  57.6  & 67.8 \\
    Fin-R1-SFT & 7B & 73.0 & 81.0 & 76.0 & 68.0 & 61.4 & 71.9 \\
    \textbf{Fin-R1} & 7B & \textbf{76.0} & \textbf{85.0} & 81.0 & \textbf{71.0} & \textbf{62.9} & \textbf{75.2} \\
    \bottomrule[2pt]
    \end{tabular}
}
\centering
\end{table*}
Specifically, we compare Fin-RL against two variants: (1) Fin-R1-SFT, trained only with SFT; and (2) Fin-R1-Zero, trained exclusively using the GRPO algorithm. For reference, we also include the base model Qwen2.5-7B-Instruct and report the results in Table \ref{tab:model_ablation}. It can be seen that  directly applying GRPO improves the base model’s capabilities, but the gains are modest. A closer look at the model output (see Figure \ref{ablationstudy1}) reveals that using GRPO alone for RL training often produces incoherent outputs.
To the contrary, Fin-R1-SFT achieves much higher scores across most datasets. the value of high-quality reasoning trace data in Fin-R1: SFT leverages these traces to guide model outputs, GRPO relies solely on final answers and ignores reasoning traces (see Figure \ref{fig:2stage}). As a result, GRPO-based training can produce incoherent outputs in pursuit of reward maximization. 
However, there is a notable gap between Fin-R1-SFT and Fin-R1, demonstrating the effectiveness of the two-step post-training in enhancing the reasoning capabilities of financial LLMs. 

Finally, to save space, we present a case study in Section~\ref{sec:case-study} of the Supplementary Material to showcase Fin-R1’s outstanding performance in financial applications.

\section{Conclusion}
\label{s4}
We introduce Fin-R1, a financial LLM, in this paper to simultaneously address three core challenges in finance: fragmented financial data, intransparent reasoning processes of existing financial LLMs and their weak business generalization capabilities. We construct Fin-R1-Data—a high-quality financial reasoning CoT dataset—and train Fin-R1 using a combination of SFT and GRPO. Our model achieves state-of-the-art performance, outperforming much larger LLMs with over one-hundred-times more parameters, and attains the best scores of 85.0 and 76.0 on the ConvFinQA and FinQA benchmark datasets, respectively. Our proposal considerably advances the application of LLMs in finance.

In the future, we will focus on advancing the field of fintech along two directions. First, we will refine Fin-R1’s architecture to accommodate financial multimodal data and deepen its application in cutting-edge areas. Second, we will promote the widespread adoption of LLMs in finance by fostering deeper integration with risk management and regulatory compliance, ultimately expanding the practical utility of Fin-R1.

\begin{spacing}{1.7}
	\setlength{\bibsep}{0.8pt}
		\bibliographystyle{apalike}
		\bibliography{references.bib}
\end{spacing}

\clearpage
\appendix

THIS SUPPLEMENT IS STRUCTURED as follows. Section~\ref{sec:Datasets} systematically compiles information on the sources of the datasets used in our experiments and case studies, covering dataset categories, content descriptions, data sources, open-source status, language types, data types, and the sizes of the original and processed data; it also provides a structured table summarizing the key datasets employed in our financial AI experiments. Section~\ref{sec:Prompt of Judging}  presents prompt templates for data distillation and filtering: Subsection~\ref{sec: prompt_example_data_distillation} introduces the data-distillation prompt, which follows the official DeepSeek-R1 configuration and comprises the task description, task input, task instruction, an explicit execution directive, and normative notes, together with the corresponding figure; Subsections~\ref{sec: anw-che} and ~\ref{sec: prompt_example_reasoning_selection} outline   the answer-check prompt  and the reasoning-selection prompt, respectively. The former consists of a role definition, an input module, judgment rules for consistency assessment, and output-format requirements; the latter is designed around seven key dimensions and specifies evaluation criteria, input components, and the scoring method. Both prompts are accompanied by supporting figures. Subsection~\ref{sec: model_selection_answer_check} evaluates the impact of five prompt formats on evaluation-model performance, builds a quantitative evaluation metric system, and accordingly identifies Qwen2.5-72B-Instruct as the judge model for answer check; the experimental data and performance-comparison results both support this conclusion.  Subsection~\ref{sec: model_selection_reasoning_selection} reports supplementary experiments conducted to select the reasoning scoring model: we compare the scoring outcomes of Qwen2.5-72B-Instruct, GPT-4o, and human annotators, and visualize correlations among their score distributions using heatmaps.
Section~\ref{sec:case-study} presents a case study that intuitively demonstrates Fin-R1’s superior reasoning and response performance in financial scenarios.

\section{Details for raw datasets}
\label{sec:Datasets}

Here we provide information on the sources of the datasets used in our experiments and case studies. The table~\ref{tab:financial_data_details} systematically classifies financial data by different business scopes, including financial advanced business knowledge, financial non-reasoning business knowledge,
financial professional knowledge, and financial code, providing a comprehensive overview of the data composition in experiments.
The table~\ref{tab:financial_data_source} elaborates on each financial data source, detailing aspects like whether the data is open, the language used, the data type, and both the original and processed sizes. 

\begin{itemize}[leftmargin=*]\item \textbf{FinanceQT \citep{malik2024quanttradingxu2023}}: Focuses on generating financial code and quantitative strategy scripts for financial scenarios.\item \textbf{Finance-500K \citep{flowers2025financeinstruct}}: Covers professional financial content such as financial terminology explanation and QA on financial expertise.\item \textbf{FinanceIQ \citep{financeIQ2023}}: Used for tasks including financial terminology explanation, QA on financial expertise, and financial calculations.\item \textbf{FinPEE (Constructed by us)}: Centered on financial calculation tasks, built based on original exam questions of Shanghai University of Finance and Economics.\item \textbf{Ant-Finance \citep{alipay2023financial}}: Involves financial business tasks that do not require complex reasoning, such as content generation and compliance management.\item \textbf{FinCorpus \citep{duxiaoman2023fincorpus}}: Contains corpora related to financial domain knowledge acquisition and financial text generation.\item \textbf{FinQA \citep{chen-etal-2021-finqaa}}: Focuses on reasoning tasks for financial data, such as numerical reasoning in financial business.\item \textbf{ConvFinQA \citep{chen2022convfinqa}}: Used for QA tasks in the financial field that require context-aware reasoning.\item \textbf{TFNS \citep{anonymous2024twitter}}: Mainly supports reasoning tasks with sentiment analysis attributes, such as sentiment classification of financial news.\item \textbf{FinCUGE \citep{lu2023bbtfin}}: Involves multiple types of reasoning tasks in the financial field, including numerical analysis and causal relationship extraction.\end{itemize}

\begin{table}[h]
    \centering
    \renewcommand{\arraystretch}{2.1}
    \setlength{\tabcolsep}{6pt}
    \caption{Financial Data Category Details}
    \label{tab:financial_data_details}
    \begin{tabularx}{\textwidth}{p{3cm} >{\raggedright\arraybackslash}X p{2.5cm} p{2cm} p{1.5cm}}
        \toprule[1.5pt]
        \textbf{Data Category} & \textbf{Description} & \textbf{Source} & \textbf{Proportion} & \textbf{Number} \\
        \midrule[1.0pt]
        Financial Code & Financial Quantitative Strategy Code Generation & FinanceQT & 0.25\% & 152 \\
        \midrule[0.5pt]

        \multirow{3}{*}{\makecell[l]{Financial\\Professional\\Knowledge}} &
        \multirow{3}{*}{\makecell[l]{Covers professional financial\\ content like terminology \\explanation, expertise Q\&A,\\ and calculations}} 
        & Finance-500K & 18.80\% & 11300 \\
        & & FinanceIQ & 4.32\% & 2596 \\
        & & FinPEE & 0.30\% & 179 \\

        \midrule[0.5pt]

        \multirow{2}{*}{\makecell[l]{Financial\\Non-reasoning\\Business\\Knowledge}} & 
        \multirow{2}{*}{\makecell[l]{Content Generation in\\ Financial Business,\\ Regulatory Compliance,\\ Financial Knowledge}} 
        & Ant-Finance & 2.58\% & 1548 \\
        & & FinCorpus & 48.74\% & 29288 \\

        \midrule[0.5pt]

        \multirow{4}{*}{\makecell[l]{Financial\\Reasoning\\Business\\Knowledge}} & 
        \multirow{4}{*}{\makecell[l]{Numerical Reasoning on \\Financial Data, Financial\\ News Sentiment Classification,\\ Financial News Classification,\\ Financial Causal \\Relationship Extraction}} 
        & FinQA & 4.91\% & 2948 \\
        & & ConvFinQA & 12.70\% & 7629 \\
        & & TFNS & 4.08\% & 2451 \\
        & & FinCUGE & 3.33\% & 2000 \\

        \bottomrule[2.0pt]
    \end{tabularx}
\end{table}

\begin{table}[h]
\centering
\caption{Financial Data Source}
\begin{tabular}{>{\centering\arraybackslash}p{2.5cm} >{\centering\arraybackslash}p{1.5cm} >{\centering\arraybackslash}p{2cm} >{\centering\arraybackslash}p{2.5cm} >{\centering\arraybackslash}p{2.6cm} >{\centering\arraybackslash}p{3cm}}
\toprule
\textbf{Source} & \textbf{Open} & \textbf{Language} & \textbf{Data Type} & \textbf{Original Size} & \textbf{Processed Size} \\
\midrule
FinanceQT & Open & English & Choice Qs & 0.39k & 0.15k \\
Finance-500K & Open & English & QA Pairs & 518.19k & 11.30k \\
FinanceIQ & Open & Chinese & Choice Qs & 7.17k & 2.60k \\
FinPEE & Closed & Chinese & Calculation & 0.35k & 0.18k \\
Ant-Finance & Open & Chinese & Choice/T-F & 8.45k & 1.55k \\
FinCorpus & Open & Chinese & Corpus & 235.21k & 29.29k \\
FinQA & Open & English & QA Pairs & 8.28k & 2.95k \\
ConvFinQA & Open & English & QA Pairs & 14.12k & 7.63k \\
TFNS & Open & English & Sentiment & 11.93k & 2.45k \\
FinCUGE & Open & Chinese & Multi-task & 66.67k & 2.00k \\
\bottomrule
\end{tabular}
\label{tab:financial_data_source}
\end{table}

\clearpage
\section{The Prompt of Data Construction}
\label{sec:Prompt of Judging}

In this section, we introduce the prompt templates for data distillation and filtering. 

\subsection{The Prompt of Data Distillation}
\label{sec: prompt_example_data_distillation}

\begin{figure}[h!]
        \centering
    \includegraphics[width=\textwidth, height=0.25\textheight, keepaspectratio=false]{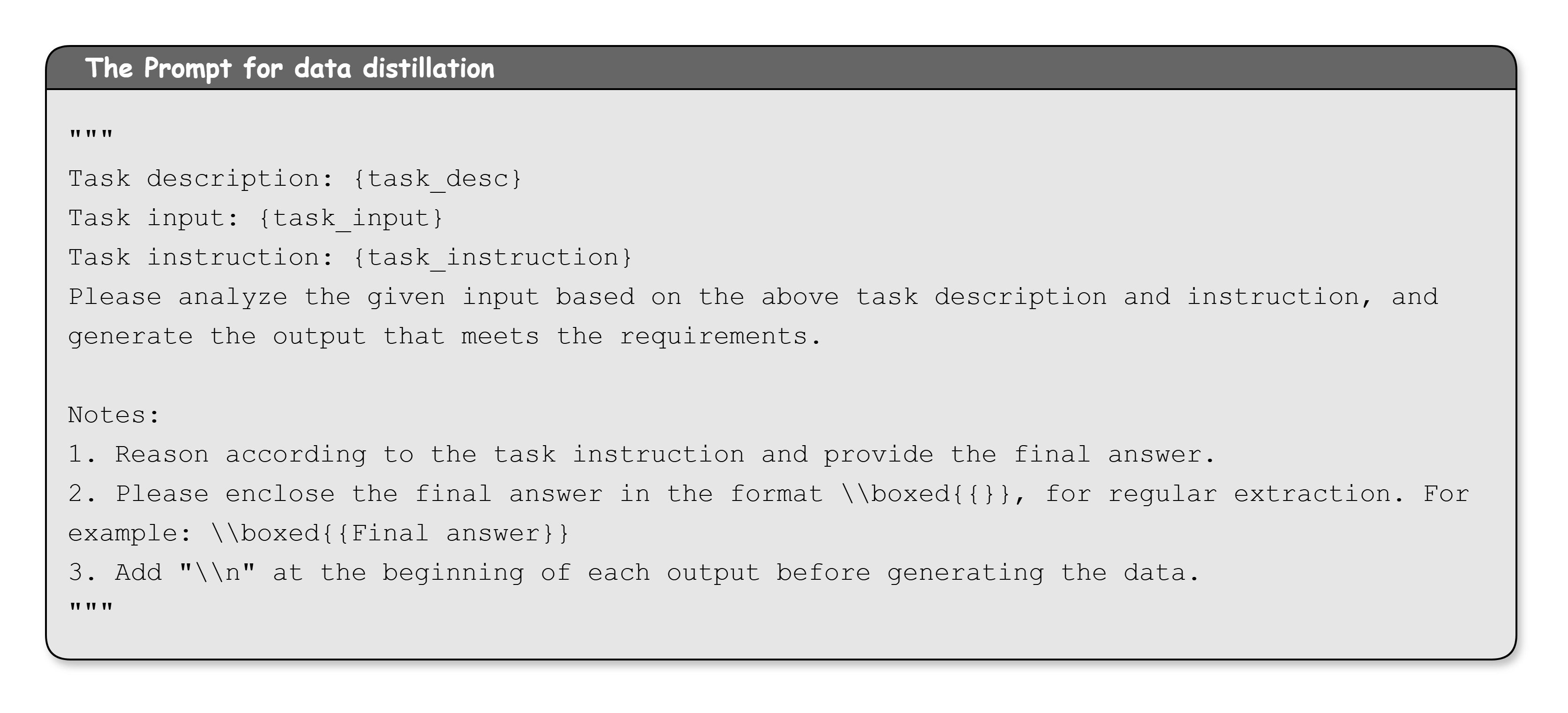} 
        \caption{The prompt of data distillation that we used.}
        \label{fig: Data_Distill_prompt}
 \end{figure}

For the data distillation step in the pipeline of data construction for Fin-R1-Data, we refer to the official prompt setting of DeepSeek-R1 and construct the prompt presented in Figure~\ref{fig: Data_Distill_prompt}. The prompt is consisted of three key components. First, it specifies the task description, input, and instruction, which define the problem, the data to be processed, and the required operation. Second, it includes an explicit execution directive (“Please analyze... and generate...”), which guides the model to align its reasoning with the task requirements. Third, it provides normative notes, requiring the model to reason according to the instruction.

\subsection{The Prompt of Answer Check}\label{sec: anw-che}
 
\begin{figure}[h!]
    \centering
    \includegraphics[width=\textwidth, height=0.6\textheight, keepaspectratio=false]{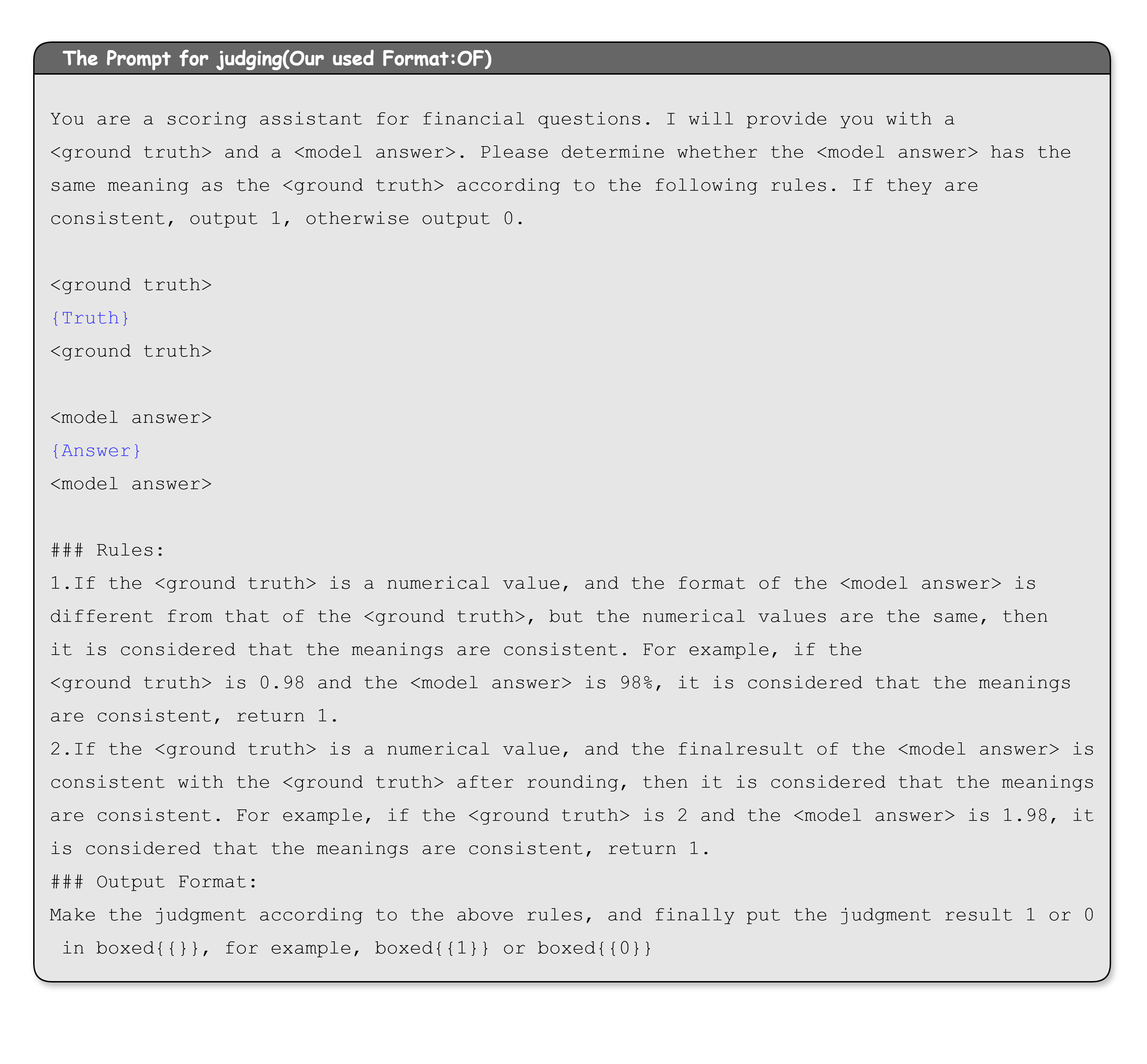} 
    \caption{The prompt for answer check that we used.}
    \label{fig: OF_prompt}
\end{figure}

 For answer check in data filtering step, the prompt presented in Figure~\ref{fig: OF_prompt} is structured into four main parts: an initial role definition that frames the model as a scoring assistant, an input section providing both the ground truth and the model answer, a set of rules that specify how numerical equivalence and rounding should be judged for consistency, and an output format requirement that constrains the response to a binary decision. 
 
\subsection{The Prompt of Reasoning Selection}\label{sec: prompt_example_reasoning_selection}
 
\begin{figure}[h!] 
        \centering 
    \includegraphics[width=\textwidth, height=0.6\textheight, keepaspectratio=false]{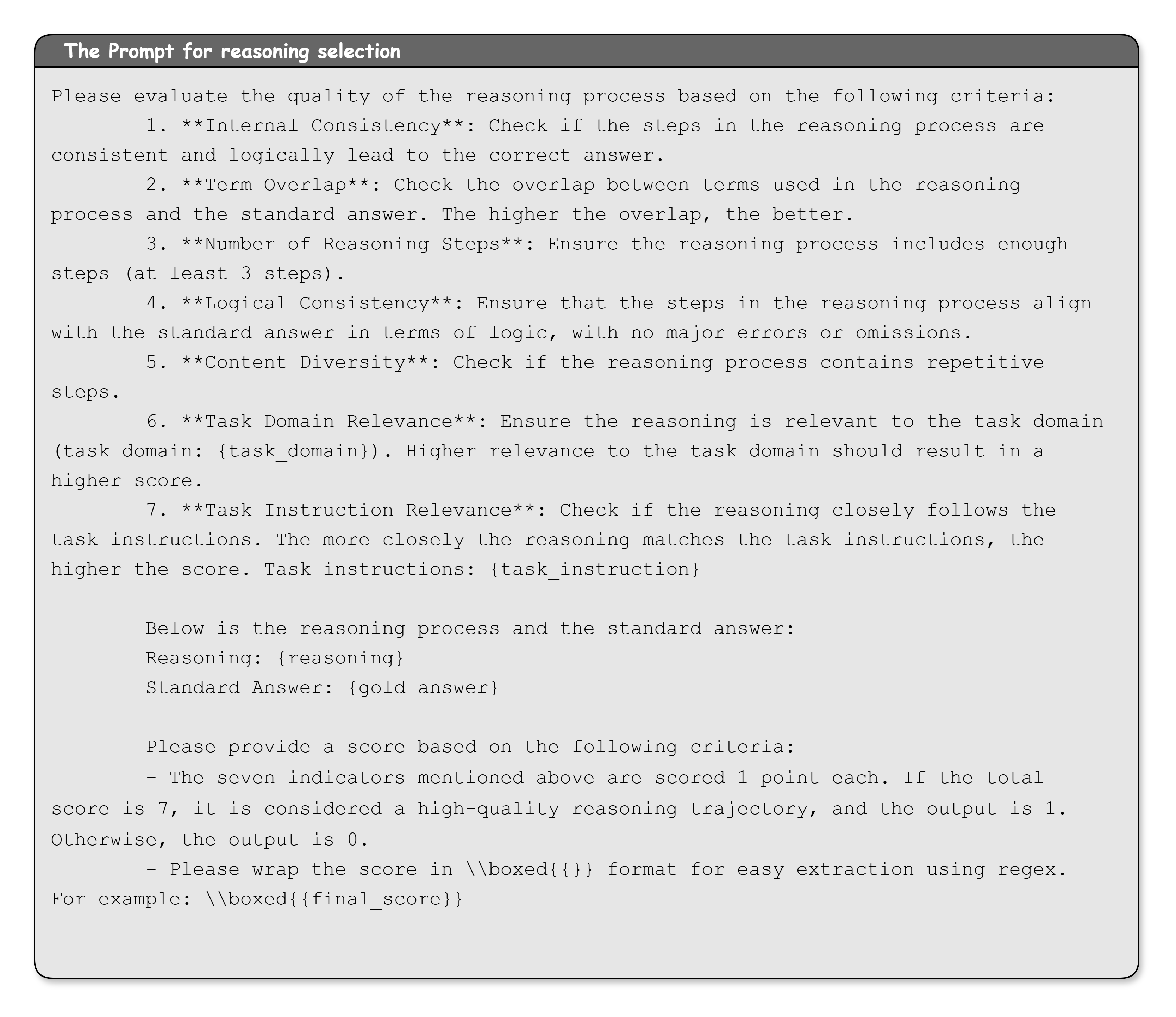} 
        \caption{The prompt for reasoning selection we used.} 
        \label{fig: reasoning_selection_prompt} 
  \end{figure}

 For reasoning selection in data filtering step, the prompt template presented in Figure~\ref{fig: reasoning_selection_prompt} is designed based $7$ key dimensions \citep{xie2024finnlpa} and is consisted of the evaluation criteria that structure how the reasoning process should be assessed; the inputs, the reasoning process and the standard answer, against which the evaluation is carried out; the description of scoring method, where each criterion is worth one point and the final decision is binary (1 for high-quality reasoning if all seven points are met, otherwise 0), with the score required for standardized extraction.

\section{Model Selection for Data Construction}

In this section, we provide the supplementary experiments to illustrate the model selection process of data construction.

\subsection{Model Selection for Answer Check}
\label{sec: model_selection_answer_check}

\begin{figure}[htbp] 
        \centering 
    \includegraphics[width=\textwidth, height=0.6\textheight, keepaspectratio=false]{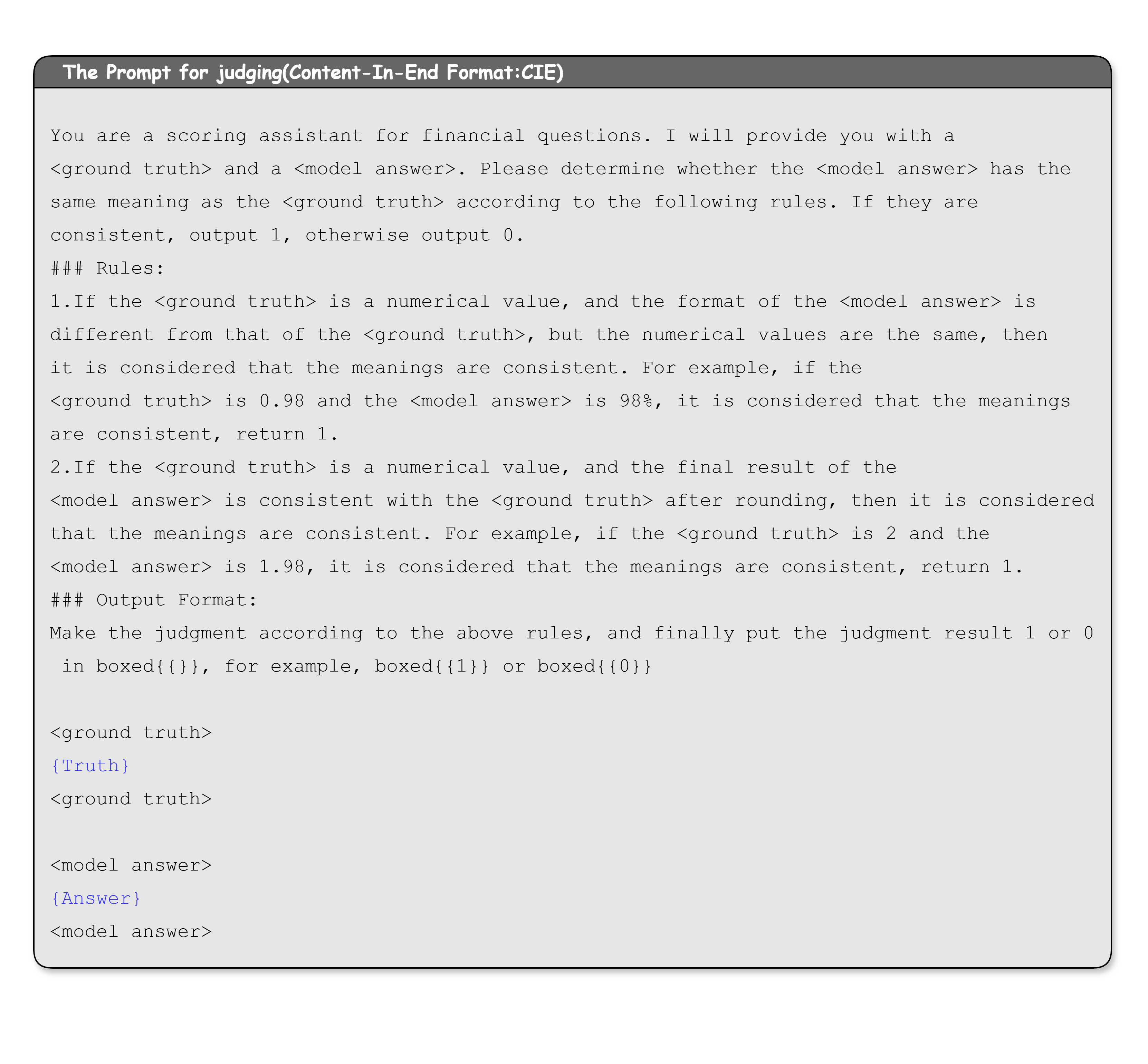} 
        \caption{The prompt for answer check with input at the end.} 
        \label{fig:CIE_prompt} 
    \end{figure}

\begin{figure}[htbp] 
        \centering 
    \includegraphics[width=\textwidth, height=0.6\textheight, keepaspectratio=false]{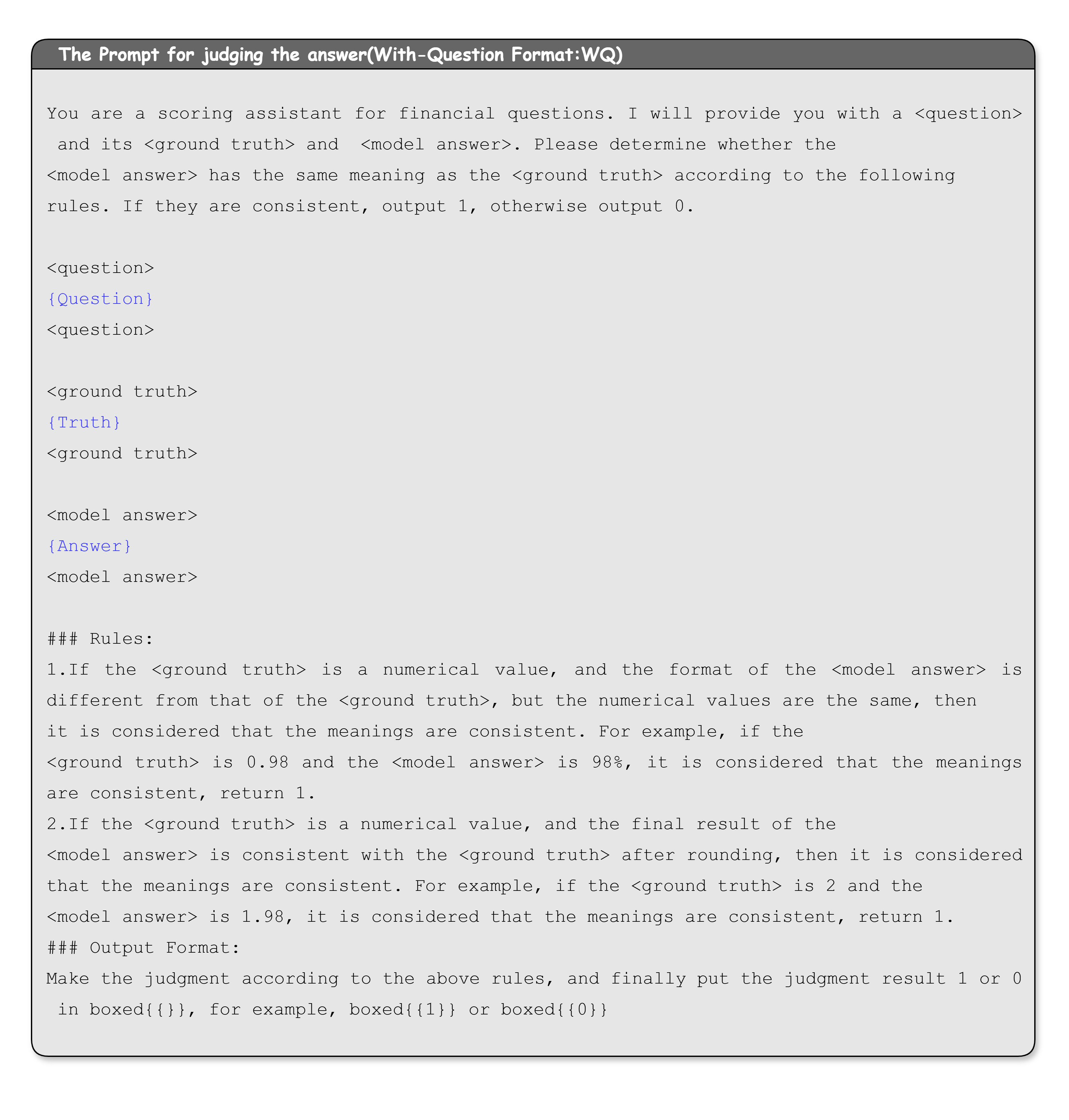} 
        \caption{The prompt for answer check with question in input.} 
        \label{fig:WQ_prompt} 
    \end{figure}

\begin{figure}[htbp] 
        \centering 
    \includegraphics[width=\textwidth, height=0.6\textheight, keepaspectratio=false]{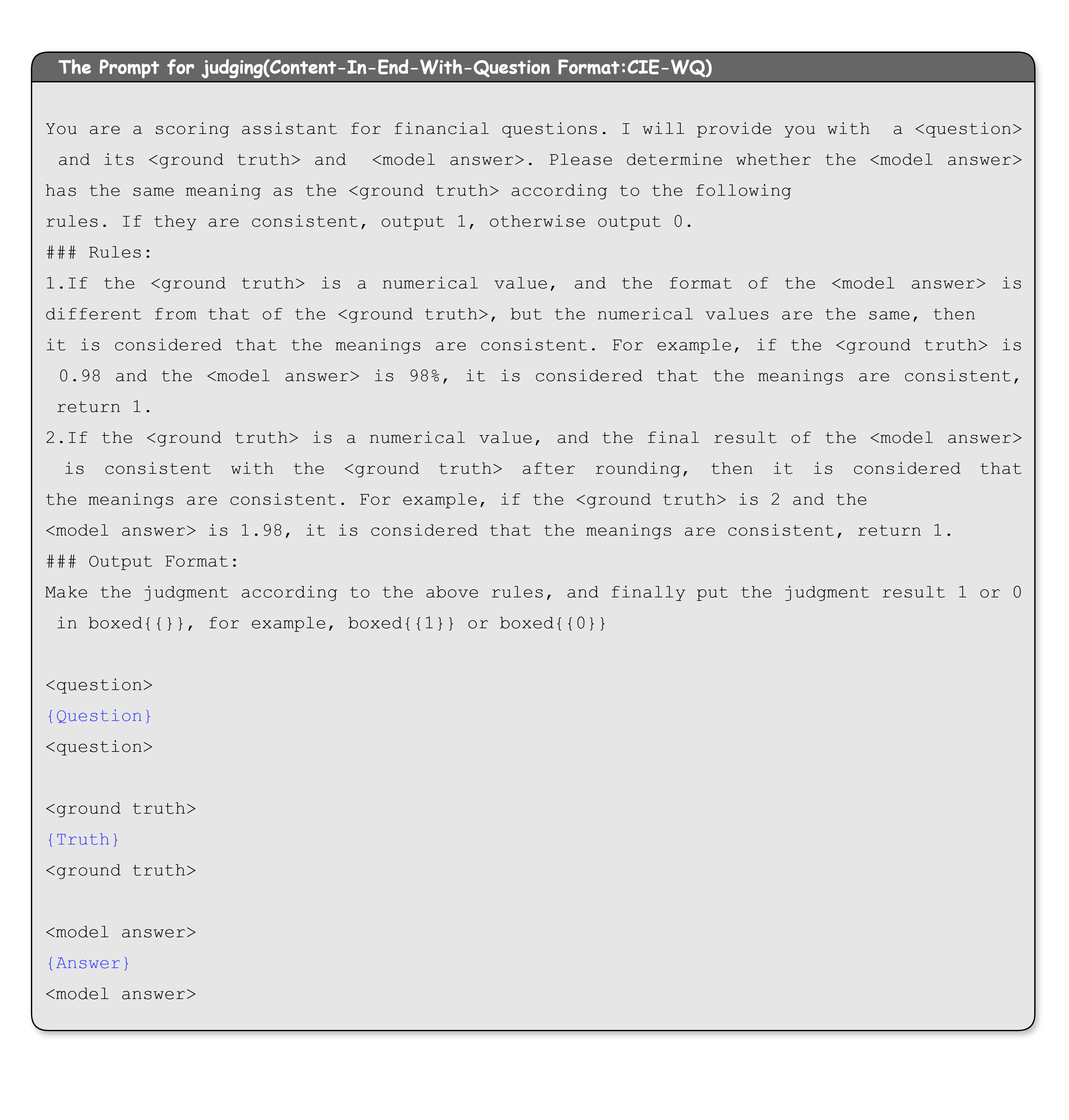} 
        \caption{The prompt for answer check with input containing question at the end.} 
        \label{fig:CIE-WQ_prompt} 
    \end{figure}

\begin{figure}[htbp] 
        \centering 
    \includegraphics[width=\textwidth, height=0.6\textheight, keepaspectratio=false]{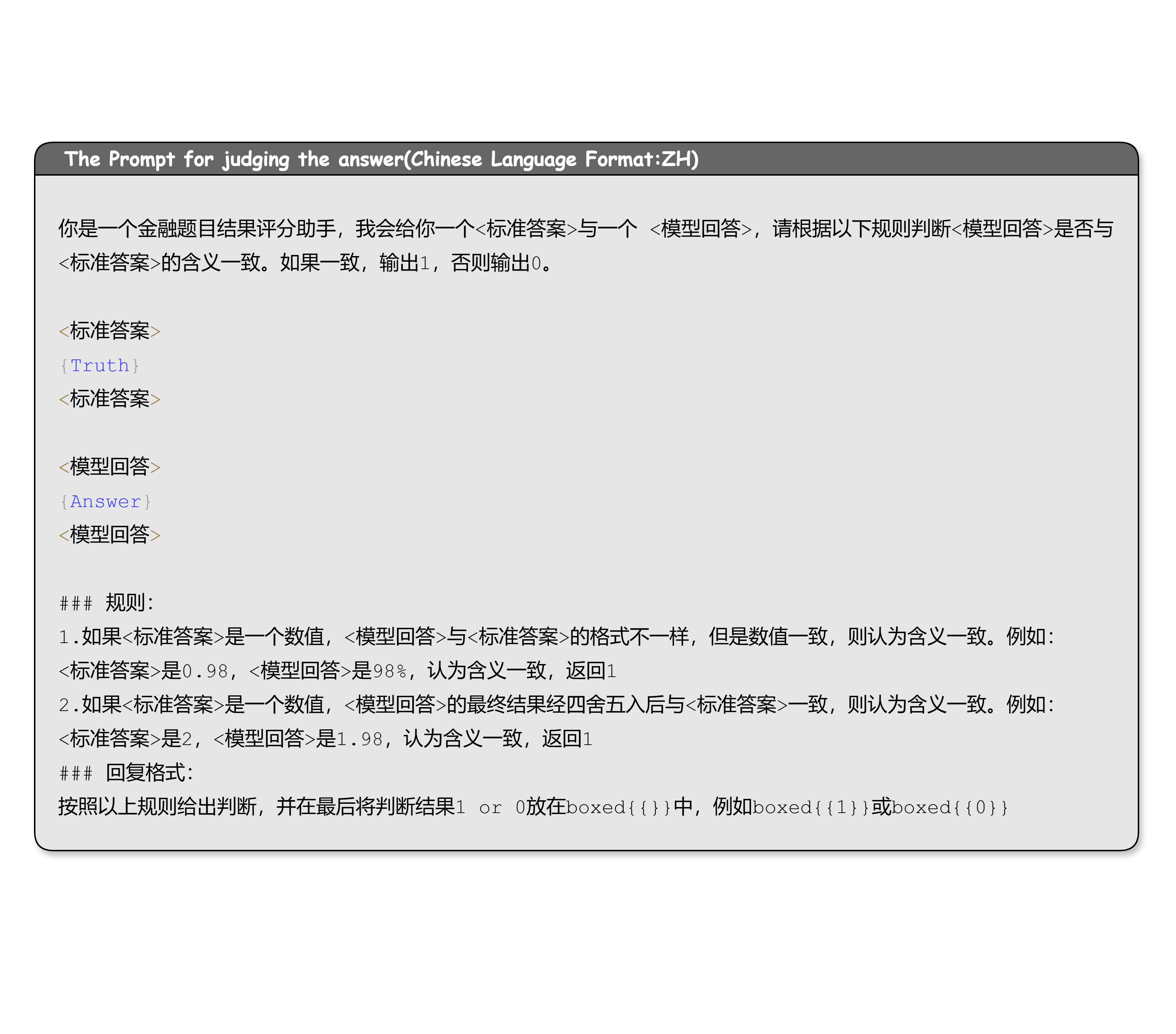} 
        \caption{The Chinese prompt for answer check.} 
        \label{fig:ZH_prompt} 
    \end{figure}

In the research on answer verification tasks based on LLM-as-Judge, we reveals that although the surface task format appears relatively simple (i.e. determining binary output 1 or 0 based on consistency between model-generated answers and reference answers), different prompt wording strategies significantly influence the performance of evaluation models. To quantitatively analyze this phenomenon, we design $5$ different prompt templates presented in Figure~\ref{fig: OF_prompt}, \ref{fig:CIE_prompt}, \ref{fig:WQ_prompt}, \ref{fig:CIE-WQ_prompt}, \ref{fig:ZH_prompt}, our used format (OF), the format where the content to be judged is at the end (CIE), the format with the original question passed in (WQ), the format with the original question passed in and the question-and-answer content placed at the end (CIE-WQ), and the Chinese format (ZH). We randomly select 100 sample instances from the FinQA dataset and conduct five repeated experiments for each prompt strategy to assess result stability. This process yielded 500 comparative results per prompt group, with model performance evaluated through consistency analysis against human-annotated results. Experiments were conducted using both GPT-4o and Qwen2.5-72B-Instruct.

To systematically evaluate the impact of different prompting strategies on evaluation model performance, this study established a quantitative evaluation metrics system comprising two core indicators:
\begin{itemize}
\item \textbf{Classification Inaccuracy:} defined as the proportion of samples where model judgments disagree with human annotations.
\item \textbf{Format Irregularity:} reflecting the degree to which model outputs fail to strictly adhere to binary constraints (0/1). 
\end{itemize}

Through statistical analysis of 500 comparative results under each prompting strategy, the performance comparison data are shown in Table \ref {tab:judge_table}. 
\begin{table}[htbp]
    \centering
    \begin{tabular}{lcccc}
        \specialrule{1.2pt}{0pt}{0pt}
        & \multicolumn{2}{c}{\textbf{Inaccuracy}} & \multicolumn{2}{c}{\textbf{Irregularity}} \\
        \cmidrule(r){2-3} \cmidrule(r){4-5}
        \textbf{Format} & GPT-4o & Qwen2.5-72B-Instruct & GPT-4o & Qwen2.5-72B-Instruct \\
        \midrule
        OF & 2.8\% & 0.4\% & 0.8\% & 0.0\% \\
        CIE & 2.0\% & 2.0\% & 0.0\% & 0.0\% \\
        WQ & 6.0\% & 8.0\% & 3.6\% & 3.2\% \\
        CIE-WQ & 4.8\% & 9.6\% & 1.6\% & 3.2\% \\
        ZH & 5.2\% & 1.6\% & 0.0\% & 0.0\% \\
        \specialrule{1.2pt}{0pt}{0pt}
    \end{tabular}
    \caption{Comparison of GPT-4o and Qwen2.5-72B-Instruct on answer judgment inaccuracy and irregularity across different prompt formats.}
    \label{tab:judge_table}
\end{table}
The systematic analysis based on experimental data reveals that different prompt strategies significantly influence the performance of evaluation models. We analyze the results as follows: 
\begin{itemize}
\item Text positioning strategies demonstrate model-specific differences. GPT-4o shows stable performance under the CIE strategy when reference answers are post-positioned, with an inaccuracy rate of 2.0\%, while Qwen2.5-72B-Instruct exhibits superior adaptation to the rule-preceding OF strategy, achieving an extremely high accuracy of 99.6\%.

\item Although incorporating original questions as contextual information theoretically enhances semantic comprehension, it substantially increases the format deviation rates (Irregularity). Under the WQ strategy, GPT-4o and Qwen2.5-72B-Instruct exhibit 3.6\% and 3.2\% Irregularity respectively. Manual verification identifies that format deviations predominantly occur in long-text samples, potentially due to input sequence elongation inducing model hallucinations (e.g., Qwen2.5-72B-Instruct's classification error rate under WQ strategy surges from baseline 0.4\% to 8.0\%). 
\item Cross-lingual testing indicates that Chinese prompts (ZH), while partially ensuring format compliance, yield significantly higher classification errors than optimal English strategies due to the English evaluation context. Compared with GPT-4o, Qwen2.5-72B-Instruct demonstrates better Chinese prompt adaptability. 
\end{itemize}

Based on the above analyses, we ultimately select Qwen2.5-72B-Instruct as the judge model to check the answer generated by data distillation, which achieves both high accuracy and regularity with prompt template we used.

\subsection{Model Selection for Reasoning Selection}
\label{sec: model_selection_reasoning_selection}

To compare the scoring outcomes between human annotators and language models, we conducted supplementary experiments. Specifically, we randomly selected 20 data points from the dataset filtered in the initial preprocessing step and evaluated their reasoning performance using Qwen2.5-72B-Instruct and GPT-4o. The evaluation followed seven predefined judgment criteria. Each data point received a score of 1 if its reasoning satisfied a given criterion and 0 otherwise. The total score for each data point was obtained by summing across all criteria, resulting in a range from 0 (minimum) to 7 (maximum). Given the scoring framework, we effectively employed a binary scoring approach (0/1) at the criterion level.

\begin{figure}[h]
    \centering
    \begin{minipage}[t]{0.45\textwidth}
        \centering
        \includegraphics[width=\textwidth]{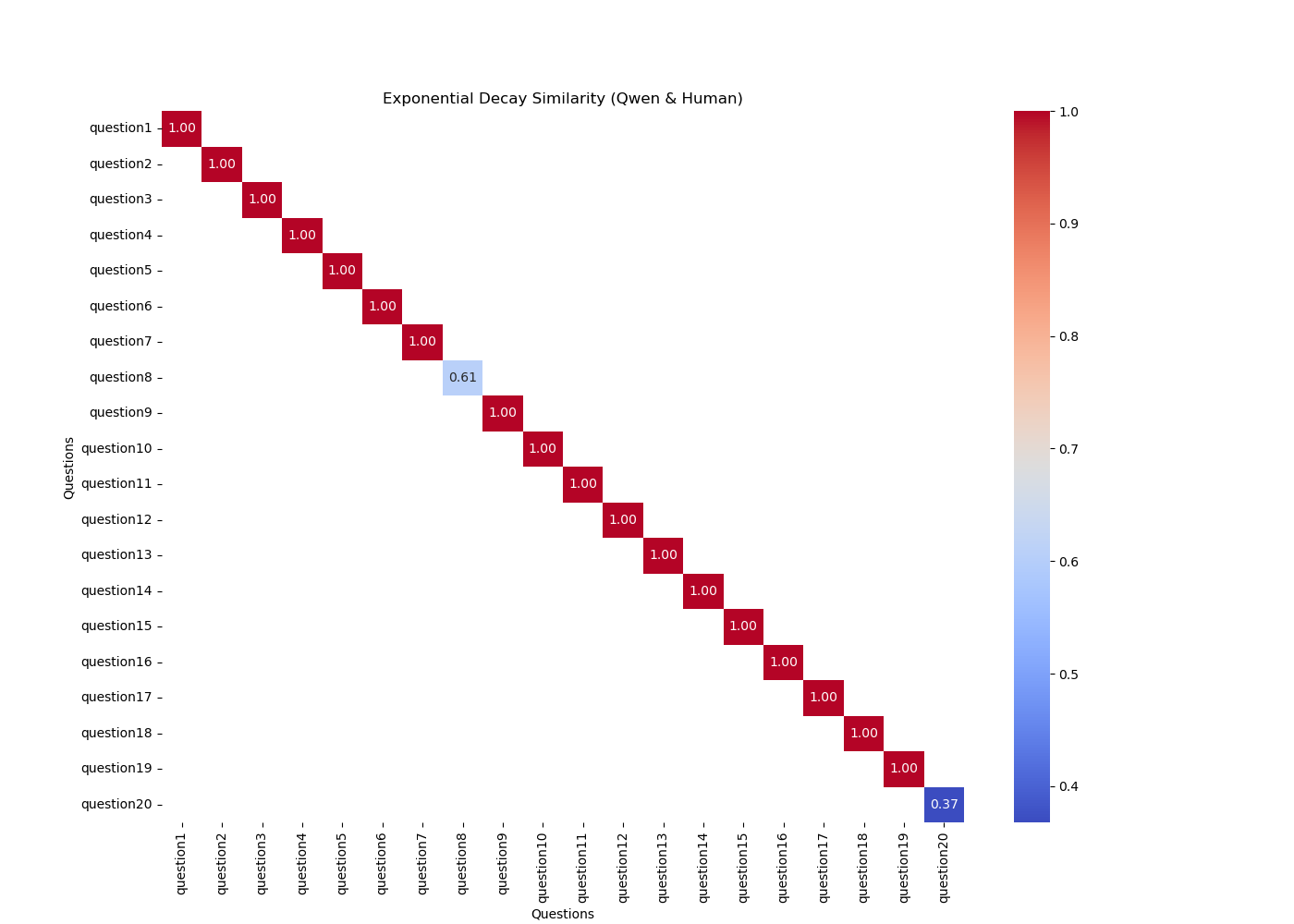}
        \subcaption{Qwen2.5-72B-Instruct vs. Human}
        \label{qweneval}
    \end{minipage}
    \hfill
    \begin{minipage}[t]{0.45\textwidth}
        \centering
        \includegraphics[width=\textwidth]{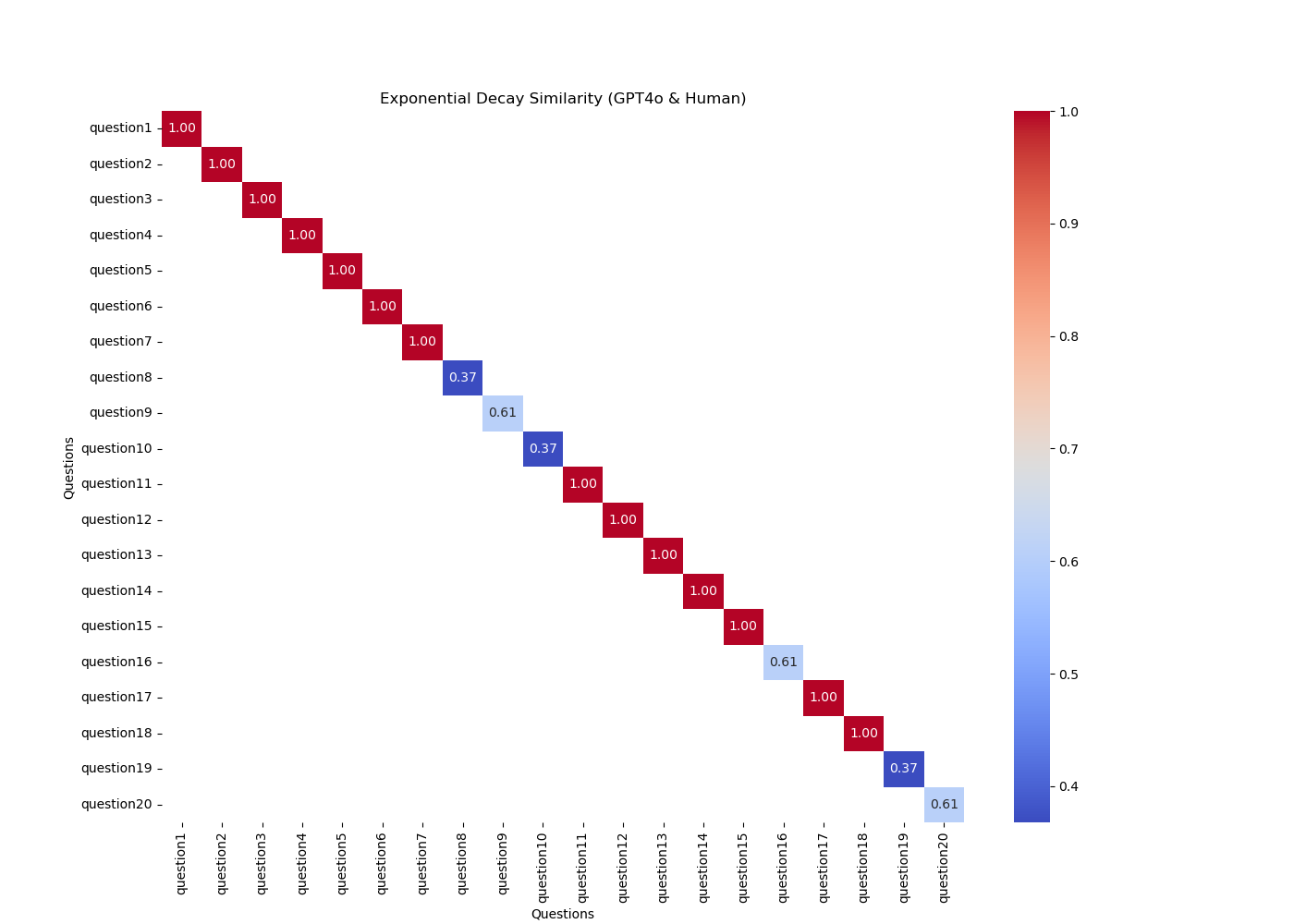}
        \subcaption{GPT-4o vs. Human}
        \label{gpteval}
    \end{minipage}
    \caption{Heatmap comparison of reasoning scores between LLMs and human annotators. Figure \ref{qweneval} and \ref{gpteval} represent the correlation between the scores of Qwen2.5-72B-Instruct and GPT-4o with human scores.}
\label{fig:Heatmap}
\end{figure}

To establish a reference baseline, human annotators independently scored the reasoning for the same data points. We then visualized the correlation between the scoring distributions of Qwen2.5-72B-Instruct, GPT-4o, and human annotations using heatmaps (see Figure \ref{fig:Heatmap}) to assess their alignment and discrepancies. The results show that Qwen2.5-72B-Instruct exhibits high concordance with human annotations, with most questions having a correlation score of 1, and only minor deviations in a few cases. In contrast, GPT-4o shows larger discrepancies, indicating lower alignment with human judgments. Based on these findings, we ultimately selected Qwen2.5-72B-Instruct as the scoring model for reasoning selection.

\section{Case Study}\label{sec:case-study}

\begin{figure}[htbp]
    \centering 
    \begin{subfigure}[b]{\textwidth} 
        \centering
        \includegraphics[width=0.6\textwidth]{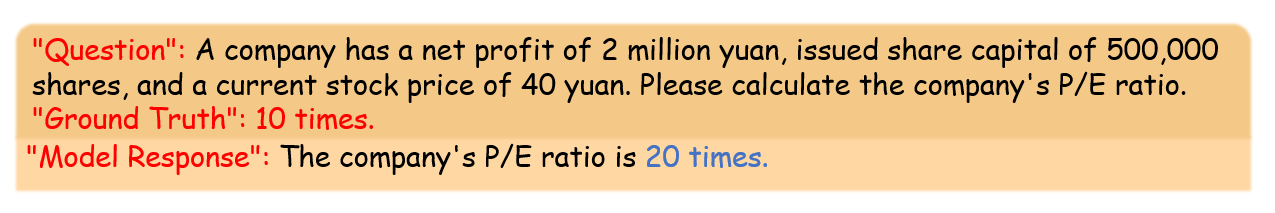} 
        \caption{Qwen2.5-7B-Instruct output without a prompt}
        \label{casestudy1}
    \end{subfigure}
    
    \vspace{10pt} 
    
    \begin{subfigure}[b]{\textwidth}
        \centering
        \includegraphics[width=0.6\textwidth]{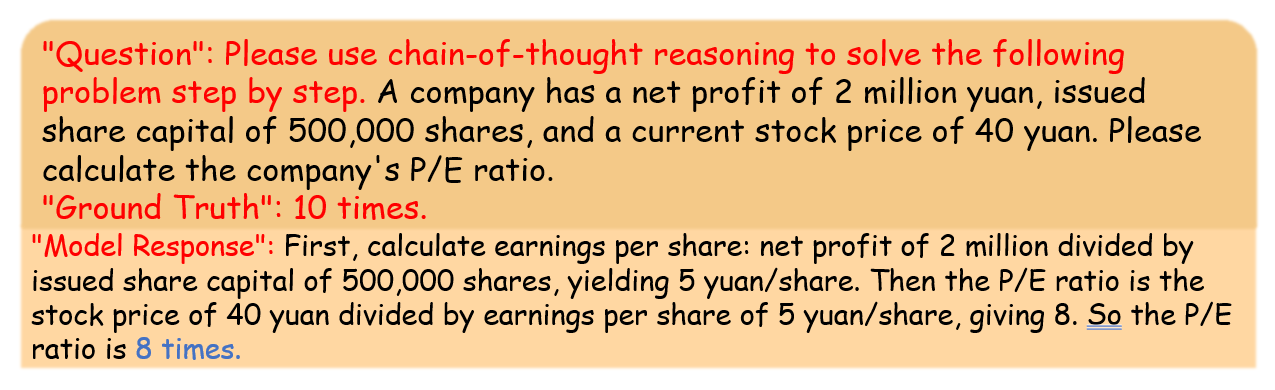}
        \caption{Qwen2.5-7B-Instruct output with a prompt}
        \label{casestudy2}
    \end{subfigure}
    \begin{subfigure}[b]{\textwidth}
        \centering
        \includegraphics[width=0.6\textwidth]{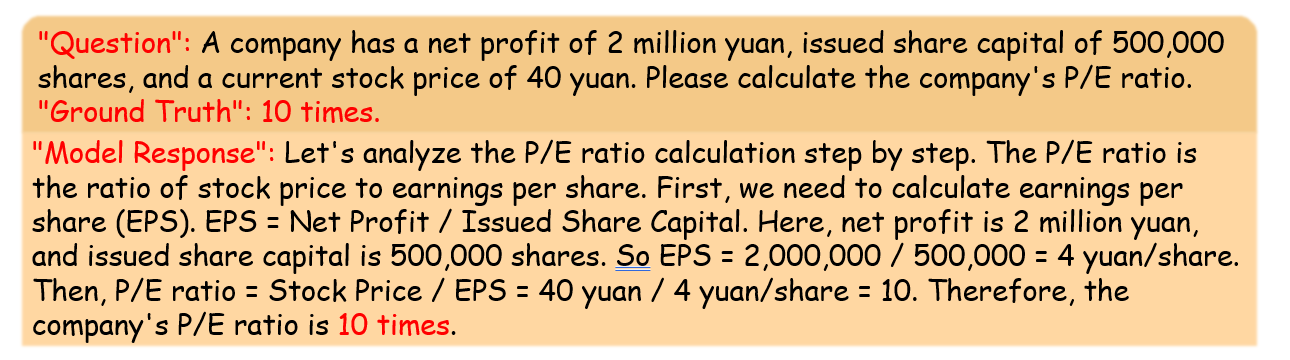}
        \caption{Fin-R1 output without a prompt}
        \label{casestudy3}
    \end{subfigure}
    
    \caption{The specific examples from the case study, where Figure~\ref{casestudy1} and Figure~\ref{casestudy2} are the outputs of Qwen2.5-7B-Instruct without and with a prompt, respectively, and Figure~\ref{casestudy3} is the high-quality output of Fin-R1. The incorrect answers are marked in blue, and the correct answers are marked in red, consistent with the ground truth.}
    \label{casestudy}
\end{figure}

Based on the actual performance of Fin-R1 in the financial domain, we conducted a case study to more intuitively demonstrate the excellent performance of Fin-R1 in financial scenarios. Figure \ref{casestudy} shows an example of the actual interactive output of both Fin-R1 and the base model Qwen2.5-7B-Instruct in a financial securities investment scenario. From the outputs of the two models, it is evident that when the prompt does not specify a particular task output format, Qwen2.5-7B-Instruct fails to respond in a structured "thought-first, then-answer" format, which is essential for meeting the requirements of reasoning tasks in the financial domain. Even when the task output format is specified, Qwen2.5-7B-Instruct's performance remains suboptimal. In contrast, Fin-R1 outperforms Qwen2.5-7B-Instruct in both logical reasoning and accuracy of its responses. Not only does Fin-R1 provide high-quality reasoning processes, but its answers are also correct and more focused.

\end{document}